%% file: NGLO.tex
\documentclass[sigconf]{acmart}

\usepackage{bm}
\usepackage{multirow}
\usepackage{enumitem}
\usepackage{colortbl}
\usepackage{booktabs}    % \cmidrule
\usepackage{pdfpages}

\definecolor{mygray}{gray}{.85}

\newcommand{\norm}[1]{\left\Vert#1\right\Vert}
\newcommand{\etal}{\textit{et al}.}
\newcommand{\ie}{\textit{i}.\textit{e}.}
\newcommand{\eg}{\textit{e}.\textit{g}.}
\let\sss\scriptscriptstyle

\usepackage[ruled]{algorithm2e} % For algorithms

\SetAlFnt{\small}
\SetAlCapFnt{\small}
\SetAlCapNameFnt{\small}
\SetAlCapHSkip{0pt}

\copyrightyear{2023}
\acmYear{2023}
\setcopyright{acmlicensed}\acmConference[SA Conference Papers '23]{SIGGRAPH Asia 2023 Conference Papers}{December 12--15, 2023}{Sydney, NSW, Australia}
\acmBooktitle{SIGGRAPH Asia 2023 Conference Papers (SA Conference Papers '23), December 12--15, 2023, Sydney, NSW, Australia}
\acmPrice{15.00}
\acmDOI{10.1145/3610548.3618253}
\acmISBN{979-8-4007-0315-7/23/12}

\begin{document}
% Title portion
\title{Neural Gradient Learning and Optimization for Oriented Point Normal Estimation}

% Authors
\author{Qing Li}
% \orcid{0009-0009-8371-4319}
\affiliation{
 \institution{School of Software, Tsinghua University}
 \city{Beijing}
 \country{China}}
\email{leoqli@tsinghua.edu.cn}

\author{Huifang Feng}
% \orcid{0000-0002-6874-698X}
\affiliation{
 \institution{School of Informatics, Xiamen University}
 \city{Xiamen}
 \country{China}}
\email{fenghuifang@stu.xmu.edu.cn}

\author{Kanle Shi}
% \orcid{0000-0001-5865-6078}
\affiliation{
 \institution{Kuaishou Technology}
 \city{Beijing}
 \country{China}}
\email{shikanle@kuaishou.com}

\author{Yi Fang}
% \orcid{0000-0001-9427-3883}
\affiliation{
 \institution{Center for Artificial Intelligence and Robotics, New York University}
 \city{Abu Dhabi}
 \country{UAE}}
\email{yfang@nyu.edu}

\author{Yu-Shen Liu}
\authornote{The corresponding author is Yu-Shen Liu.}
% \orcid{0000-0001-7305-1915}
\affiliation{
 \institution{School of Software, Tsinghua University}
 \city{Beijing}
 \country{China}}
\email{liuyushen@tsinghua.edu.cn}

\author{Zhizhong Han}
% \orcid{0000-0001-9540-9973}
\affiliation{
 \institution{Department of Computer Science, Wayne State University}
 \city{Detroit}
 \country{USA}}
\email{h312h@wayne.edu}

\begin{abstract}
  We propose Neural Gradient Learning (NGL), a deep learning approach to learn gradient vectors with consistent orientation from 3D point clouds for normal estimation.
  It has excellent gradient approximation properties for the underlying geometry of the data.
  We utilize a simple neural network to parameterize the objective function to produce gradients at points using a global implicit representation.
  However, the derived gradients usually drift away from the ground-truth oriented normals due to the lack of local detail descriptions.
  Therefore, we introduce Gradient Vector Optimization (GVO) to learn an angular distance field based on local plane geometry to refine the coarse gradient vectors.
  Finally, we formulate our method with a two-phase pipeline of coarse estimation followed by refinement.
  Moreover, we integrate two weighting functions, i.e., anisotropic kernel and inlier score, into the optimization to improve the robust and detail-preserving performance.
  Our method efficiently conducts global gradient approximation while achieving better accuracy and generalization ability of local feature description.
  This leads to a state-of-the-art normal estimator that is robust to noise, outliers and point density variations.
  Extensive evaluations show that our method outperforms previous works in both unoriented and oriented normal estimation on widely used benchmarks.
  The source code and pre-trained models are available at \textcolor{red}{\href{https://github.com/LeoQLi/NGLO}{https://github.com/LeoQLi/NGLO}}.
\end{abstract}

%
% The code below should be generated by the tool at
% http://dl.acm.org/ccs.cfm
% Please copy and paste the code instead of the example below.
%



%
% End generated code
%


\maketitle

%%%%%%%%% BODY TEXT
%%% =================================================================================================
\section{Introduction}

\begin{figure}[t]
   \centering
   \includegraphics[width=\linewidth]{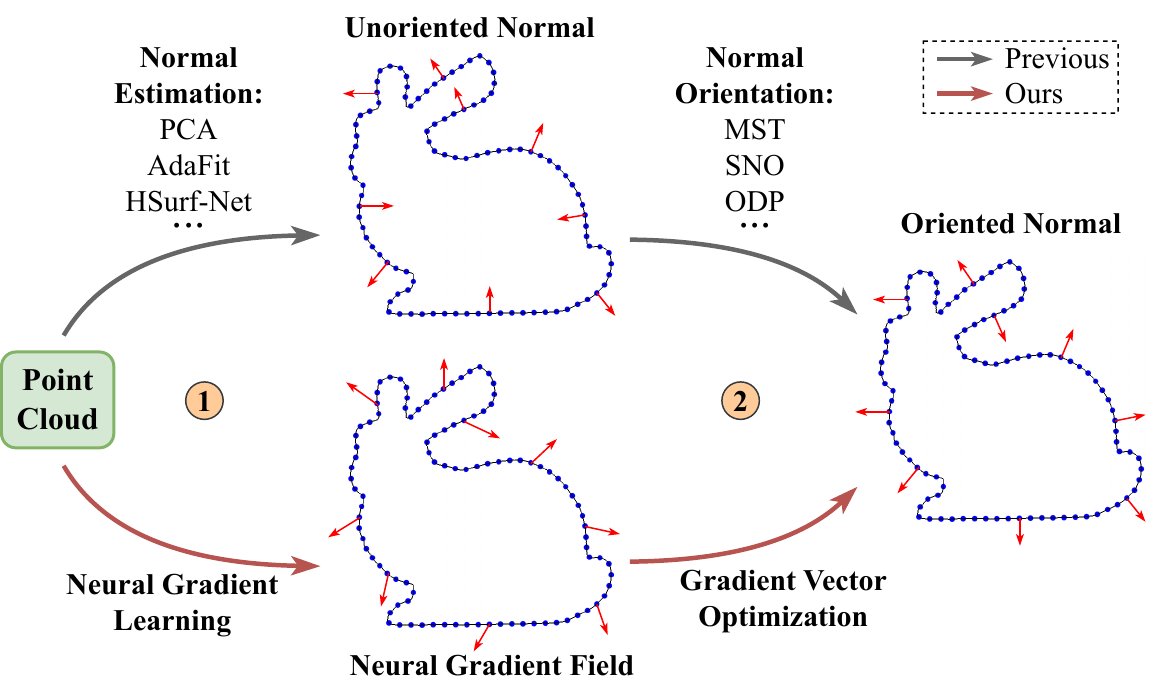}  \vspace{-0.55cm}
   \caption{
      For oriented normal estimation, previous methods usually conduct a two-stage pipeline, \ie, (1) unoriented normal estimation and (2) normal orientation, while our method achieves this through Neural Gradient Learning (NGL) and Gradient Vector Optimization (GVO).
      We introduce effective novel designs into our method that enable it to improve the SOTA results.
   }
   \label{fig:intro}
   \vspace{-0.2cm}
\end{figure}

Normal estimation is a fundamental task in computer vision and computer graphics.
Oriented normal with consistent orientation is a prerequisite for many downstream tasks, such as graphics rendering~\cite{blinn1978simulation,gouraud1971continuous,phong1975illumination} and surface reconstruction~\cite{kazhdan2005reconstruction,kazhdan2006poisson,kazhdan2013screened}.
Due to noise levels, uneven sampling densities, and various complex geometries, estimating oriented normals from 3D point clouds still remains challenging.
As shown in Fig.~\ref{fig:intro}, the paradigm of oriented normal estimation usually includes:
unoriented normal estimation that provides vectors perpendicular to the surfaces defined by local neighborhoods;
normal orientation that aligns the directions of adjacent vectors for global consistency.
Over the past few years, many excellent algorithms~\cite{lenssen2020deep,ben2020deepfit,zhu2021adafit,li2022graphfit,li2022hsurf,li2023NeAF} have been proposed for unoriented normal estimation.
However, their estimated normals are randomly oriented on both sides of the surface and cannot be directly used in downstream applications without normal orientation.
Most normal orientation approaches are based on a propagation strategy~\cite{hoppe1992surface,konig2009consistent,schertler2017towards,xu2018towards,jakob2019parallel,metzer2021orienting}.
These methods are mainly based on the assumption of smooth and clean points, and carefully tune data-specific parameters, such as the neighborhood size of the propagation.
Moreover, the issue of error propagation in the orientation process may let errors in local areas overflow into the subsequent steps.

\begin{figure}[t]
   \centering
   \includegraphics[width=\linewidth]{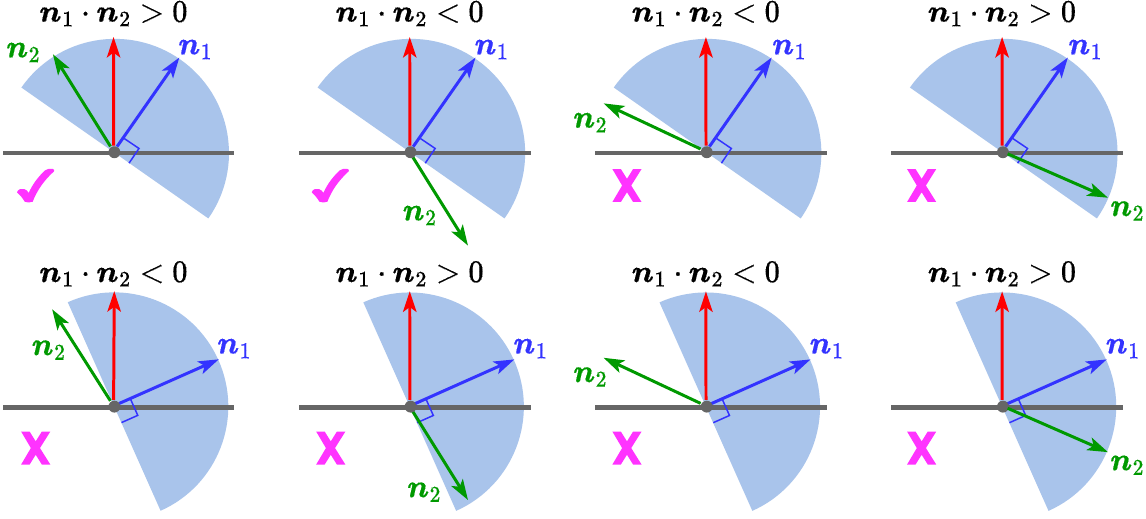}  \vspace{-0.5cm}
   \caption{
      Different cases of flipping (or not) vector $\bm{n}_2$ based on vector $\bm{n}_1$.
      Given a reference vector $\bm{n}_1$, we propagate its orientation to vector $\bm{n}_2$.
      The classic criteria is that we flip the sign of $\bm{n}_2$ if $\bm{n}_1 \!\cdot\! \bm{n}_2 \!<\! 0$.
      We can observe that there are many wrong cases according to this naive rule.
      The blue semicircle denotes the angle range, and any vector $\bm{n}_i$ within it satisfies $\bm{n}_1 \cdot \bm{n}_i \!>\! 0$.
      The surface is shown as a gray line and its ground-truth normal as a red arrow.
      We let two normal vectors be on the same point for better illustration.
      We only change $\bm{n}_2$ in each row and $\bm{n}_1$ in each column.
   }
   \label{fig:analyse}
   \vspace{-0.2cm}
\end{figure}

The two-stage architecture of existing oriented normal estimation paradigms needs to combine two independent algorithms, and requires a lot of work to tune the parameters of the two algorithms.
% solve the problems caused by code environment and data compatibility.
More importantly, the stability and effectiveness of the integrated algorithm cannot be guaranteed.
In our experiments, we evaluate the combinations of different algorithms for unoriented normal estimation and normal orientation.
A key observation is that, for the same normal orientation algorithm, integrating a better unoriented normal estimation algorithm does not lead to better orientation results.
That is, using higher precision unoriented normals does not necessarily result in more accurate oriented normals using existing propagation strategies.
In Fig.~\ref{fig:analyse}, we use a simple example to illustrate that judging whether to invert the direction of neighborhood normals based on a propagation rule will lead to unreasonable results.
The propagation strategy is affected by the direction distribution of the unoriented normal vectors.
Therefore, it is necessary to design a complete and unified pipeline for oriented normal estimation.

In a data-driven manner, the workflow of our proposed method is an inversion of the traditional pipeline (see Fig.~\ref{fig:intro}).
We start by solving normals with consistent orientation but possibly moderate accuracy, and then we further refine the normals.
We introduce \emph{Neural Gradient Learning} (NGL) and \emph{Gradient Vector Optimization} (GVO), defined by a family of loss functions that can be used with point cloud data with noise, outliers and point density variations, and efficiently produce high accurate oriented normals for each point.
Specifically, the NGL learns gradient vectors from global geometry representation, while the GVO optimizes vectors based on an insight into the local property.
A series of qualitative and quantitative evaluation experiments are conducted to demonstrate the effectiveness of the proposed method.

To summarize, our main contributions include:
\begin{itemize}[leftmargin=*]
\setlength{\itemsep}{5pt}
\setlength{\parsep}{0pt}
\setlength{\parskip}{0pt}
% \vspace{-0.2cm}
  \item A technique of neural gradient learning, which can derive gradient vectors with consistent orientations from implicit representations of point cloud data.
  \item A gradient vector optimization strategy, which learns an angular distance field based on local geometry to further optimize the gradient vectors.
  \item We report the state-of-the-art performance for both unoriented and oriented normal estimation on point clouds with noise, density variations and complex geometries.
\end{itemize}

%%% =================================================================================================
\section{Related Work}

%%% -------------------------------------------------------------------------
\subsection{Unoriented Normal Estimation}

The most widely used unoriented normal estimation method for point clouds is Principle Component Analysis (PCA)~\cite{hoppe1992surface}.
Later, PCA variants~\cite{alexa2001point,pauly2002efficient,mitra2003estimating,lange2005anisotropic,huang2009consolidation}, Voronoi-based paradigms~\cite{amenta1999surface,merigot2010voronoi,dey2006provable,alliez2007voronoi}, and methods based on complex surfaces~\cite{levin1998approximation,cazals2005estimating,guennebaud2007algebraic,aroudj2017visibility,oztireli2009feature} have been proposed to improve the performance.
These traditional methods~\cite{hoppe1992surface,cazals2005estimating} are usually based on geometric prior of point cloud data itself, and require complex preprocessing and parameter fine-tuning according to different types of data.
Recently, some studies proposed to use neural networks to directly or indirectly map high-dimensional features of point clouds into 3D normal vectors.
For example, the regression-based methods directly estimate normals from structured data~\cite{boulch2016deep,roveri2018pointpronets,lu2020deep} or unstructured point clouds~\cite{guerrero2018pcpnet,zhou2020normal,hashimoto2019normal,ben2019nesti,zhou2020geometry,zhou2022refine,li2022hsurf,li2023NeAF}.
In contrast, the surface fitting-based methods first employ a neural network to predict point weights, then they derive normal vectors through weighted plane fitting~\cite{lenssen2020deep,cao2021latent} or polynomial surface fitting~\cite{ben2020deepfit,zhu2021adafit,zhou2023improvement,zhang2022geometry,li2022graphfit} on local neighborhoods.
In our experiments, we observe that regression-based methods train models more stably and perform optimization more efficiently without coupling the fitting step used in fitting-based methods.
In contrast, our method finds the optimal point normal through a classification strategy.

%%% -------------------------------------------------------------------------
\subsection{Consistent Normal Orientation}

The normals estimated by the above methods do not preserve a consistent orientation since they only look for lines perpendicular to the surface.
Based on local consistency strategy, the pioneering work~\cite{hoppe1992surface} and its improved methods~\cite{seversky2011harmonic,wang2012variational,schertler2017towards,xu2018towards,jakob2019parallel} propagate seed point's normal orientation to its adjacent points via a Minimum Spanning Tree (MST).
More recent work~\cite{metzer2021orienting} introduces a dipole propagation strategy across the partitioned patches to achieve global consistency.
However, these methods are limited by error propagation during the orientation process.
Some other methods show that normal orientation can benefit from reconstructing surfaces from unoriented points.
They usually adopt different volumetric representation techniques, such as signed distance functions~\cite{mullen2010signing,mello2003estimating}, variational formulations~\cite{walder2005implicit,huang2019variational,alliez2007voronoi}, visibility~\cite{katz2007direct,chen2010binary}, isovalue constraints~\cite{xiao2023point}, active contours~\cite{xie2004surface} and winding-number field~\cite{xu2023globally}.
The correctly-oriented normals can be achieved from their solved representations, but their normals are not accurate in the vertical direction.
Furthermore, a few approaches~\cite{guerrero2018pcpnet,hashimoto2019normal,wang2022deep,li2023shsnet} focus on using neural networks to directly learn a general mapping from point clouds to oriented normals.
% but they have limited accuracy and their performance cannot be guaranteed across different noise levels and geometric structures.
Different from the above methods, we solve the oriented normal estimation by first determining the global orientation and then improving its direction accuracy based on local geometry.

%%% =================================================================================================
\section{Preliminary}

In general, the gradient of a real-valued function $f(x, y, z)$ in a 3D Cartesian coordinate system (also called gradient field) is given by a vector whose components are the first partial derivatives of $f$, \ie,
$\nabla f(x, y, z) \!=\! {f_x}\bm{i} + {f_y}\bm{j} + {f_z}\bm{k}$,
where $\bm{i}, \bm{j}$ and $\bm{k}$ are the standard unit vectors in the directions of the $x, y$ and $z$ coordinates, respectively.
If the function $f$ is differentiable at a point $\bm{p}$ and suppose that $\nabla f(\bm{p}) \!\neq\! 0$,
then there are two important properties of the gradient field:
(1) The maximum value of the directional derivative, \ie, the maximum rate of change of the function $f$, is defined by the magnitude of the gradient $\norm{\nabla f}$ and occurs in the direction given by $\nabla f$.
(2) The gradient vector $\nabla f$ is perpendicular to the level surface $f(\bm{p}) \!=\! 0$.

Recently, deep neural networks have been used to reconstruct surfaces from point cloud data by learning implicit functions.
These approaches represent a surface as the zero level-set of an implicit function $f$, \ie,
\begin{equation}
   \mathcal{S} = \left \{\bm{x} \in \mathbb{R}^3 ~ \vert ~ f(\bm{x}; \bm{\theta}) = 0 \right \} ,
\end{equation}
where $f\colon\mathbb{R}^{3} \!\rightarrow\! \mathbb{R} $ is a neural network with parameter $\bm{\theta}$, such as multi-layer perceptron (MLP).
Implicit function learning methods adopt either signed distance function~\cite{park2019deepsdf} or binary occupancy~\cite{mescheder2019occupancy} as the shape representation.
If the function $f$ is continuous and differentiable, the formula of normal vector (perpendicular to the surface) at a point $\bm{p}$ is $\bm{n}_{\bm{p}} \!=\! \nabla f(\bm{p})/\norm{\nabla f(\bm{p})}$, where $\|\!\cdot\!\|$ means vector norm.
Using neural networks as implicit representations of surfaces can benefit from their adaptability and approximation capability~\cite{atzmon2019controlling}.
Meanwhile, we can obtain the gradient $\nabla f$ in the back-propagation process of training $f$.

\begin{figure}[t]
   \centering
   \includegraphics[width=\linewidth]{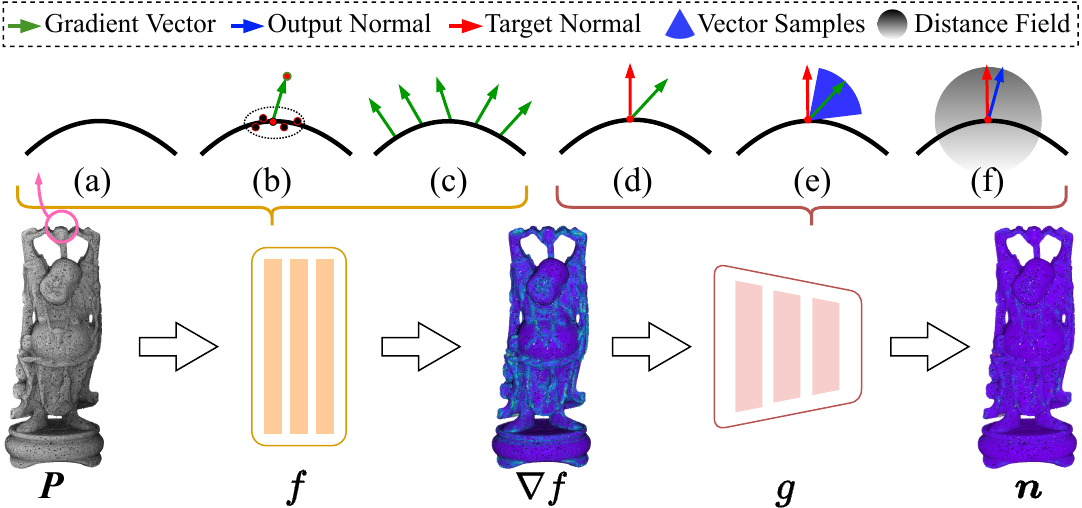}  \vspace{-0.55cm}
   \caption{
      (a-c): The neural gradient learning function $f$ takes a point cloud $\bm{P}$ as input and derives point-wise gradient $\nabla f$ within the network based on neighboring regions of the surface.
      (d-f): The gradient vector optimization function $g$ selects the optimal vector sample according to angular distance as the normal $\bm{n}$.
   }
   \label{fig:net}
   \vspace{-0.2cm}
\end{figure}

%%% =================================================================================================
\section{Method}

As shown in Fig.~\ref{fig:net}, our method consists of two parts: (1) the neural gradient learning $(\bm{P} \!\rightarrow\! f \!\rightarrow\! \nabla f)$ to estimate inaccurate but correctly-oriented gradients, and (2) the gradient vector optimization $(\nabla f \!\rightarrow\! g \!\rightarrow\! \bm{n})$ to refine the coarse gradients to obtain accurate normals, which will be introduced in the following sections.

%%% -------------------------------------------------------------------------
\subsection{Neural Gradient Learning}  \label{sec:learning}

Consider a point set $\bm{X} \!=\! \{\bm{x}_i\}_{i=1}^{M_1}$ that is sampled from raw point cloud $\bm{P}$ (possibly distorted) through certain probability distribution $\mathcal{D}$, we explore training a neural network $f$ with parameter $\bm{\theta}$ to derive the gradient during the optimization.
% First, we introduce \emph{Neural Gradient Learning} (NGL) defined by the form of a loss function
First, we introduce a loss function defined by the form of
\begin{equation} \label{eq:loss_intro}
   \mathcal{L}(\bm{\theta}) = \mathbb{E}_{\bm{x} \sim \mathcal{D}} ~\mathcal{T} \big(F(\bm{x}; \bm{\theta}), \mathcal{F}_{\bm{X}}(\bm{x}) \big) ~,
\end{equation}
where $\mathcal{T}\colon\mathbb{R}\times \mathbb{R} \rightarrow \mathbb{R}$ is a differentiable similarity function.
$F(\bm{x}; \bm{\theta})$ is the learning objective to be optimized and $\mathcal{F}_{\bm{X}}(\bm{x})$ is the distance measure with respect to $\bm{X}$.
In this work, our insight is that incorporating neural gradients in a manner similar to~\cite{atzmon2020sal,atzmon2021sald} can learn neural gradient fields with consistent orientations from various point clouds.
To this end, we add the derivative data of $f$, \ie,
\begin{equation} \label{eq:grad}
   F(\bm{x}; \bm{\theta}) = f(\bm{x}; \bm{\theta}) \cdot \bm{v} ~,
\end{equation}
where $\bm{v} \!=\! \nabla f(\bm{x}; \bm{\theta}) / \norm{\nabla f(\bm{x}; \bm{\theta})}$ is the normalized neural gradient.
Eq.~\eqref{eq:grad} incorporates an implicit representation and a gradient approximation with respect to the underlying geometry of $\bm{X}$.

We first show a special case of Eq.~\eqref{eq:loss_intro}, which is given by
\begin{equation} \label{eq:pull}
   \mathcal{L}(\bm{\theta}) = \mathbb{E}_{\bm{x} \sim \mathcal{D}} ~\mathcal{T} \big(\bm{x} - f(\bm{x}; \bm{\theta}) \cdot \bm{v},~ \bm{p} \big).
\end{equation}
Such definition of training objective has been used by surface reconstruction methods~\cite{ma2020neural,chibane2020neural} to learn signed or unsigned distance functions from noise-free data.
Recall that the gradient will be the direction in which the distance value increases the fastest.
These methods exploit this property to move a query position $\bm{x}$ by distance $f(\bm{x}; \bm{\theta})$ along or against the gradient direction $\bm{v}$ to its closest point $\bm{p}$ sampled on the manifold.
Specifically, $f(\bm{x}; \bm{\theta})$ is interpreted as a signed distance~\cite{ma2020neural} or unsigned distance~\cite{chibane2020neural}.
This way they can learn reasonable signed/unsigned distance functions from the input noise-free point clouds.
In contrast, we are not looking to learn an accurate distance field to approximate the underlying surface, but to learn a neural gradient field with a consistent orientation from a variety of data, even in the presence of noise.

% A specific example of Eq.~\eqref{eq:loss_intro} used in sign agnostic learning~\cite{atzmon2020sal,atzmon2021sald} is
% \begin{equation}
%    \mathcal{L}(\bm{\theta}) = \mathbb{E}_{\bm{x} \sim \mathcal{D}} ~\mathcal{T} \big( f(\bm{x}; \bm{\theta}) , h(\bm{x}) \big) ~,
% \end{equation}
% where $\mathcal{T}$ is taken as an unsigned similarity, \eg, $\mathcal{T}(a,b) \!=\! \big| |a|-b \big|$, and $h(\bm{y}) \!=\! \min_{\bm{x}\in\bm{X}} \norm{\bm{y}-\bm{x}}$~\cite{atzmon2020sal}.
% The key property of the sign agnostic loss is that, given proper initial weight $\bm{\theta}_0$ and an unsigned distance function $h$ to the input data, it finds a newly signed local minimum $f$ whose absolute value is similar to $h$~\cite{atzmon2021sald}.
% Thus, this signed distance function can be used as an implicit representation of the underlying surface and the zero level-set of $f$ will be a valid manifold describing the point cloud data $\bm{X}$.

Next, we will extend Eq.~\eqref{eq:loss_intro} to a more general case for neural gradient learning.
Given a point $\bm{x}$, instead of using the unsigned distance in~\cite{atzmon2020sal} or its nearest sampling point~\cite{ma2020neural,chibane2020neural}, we consider the mean vector of its neighborhood, that is
\begin{equation} \label{eq:ave_grad}
   \mathcal{F}_{\bm{X}}(\bm{x}) = \frac{1}{k}\sum_{i=1}^{k} \big(\bm{x} - \mathcal{N}^k_i(\bm{x}, \bm{P})\big ), ~ \bm{x} \in \bm{X} ~,
\end{equation}
where $\mathcal{N}^k_i(\bm{x}, \bm{P})$ denotes the $k$ nearest points of $\bm{x}$ in $\bm{P}$.
Intuitively, $\mathcal{F}_{\bm{X}}(\bm{x}) \!\in\! \mathbb{R}^3$ is a vector from the averaged point position $\bar{\bm{x}} \!=\! \sum_{i=1}^{k} \mathcal{N}^k_i(\bm{x}, \bm{P}) / k$ to $\bm{x}$.

For the similarity measure $\mathcal{T}$ of vector-valued functions, we adopt the standard Euclidean distance.
Then, the loss in Eq.~\eqref{eq:loss_intro} for \emph{Neural Gradient Learning} (NGL) has the format
\begin{equation} \label{eq:loss_grad}
   \mathcal{L}(\bm{\theta}) = \norm{f(\bm{x}; \bm{\theta}) \cdot \bm{v} - \frac{1}{k}\sum_{i=1}^{k} \big(\bm{x} - \mathcal{N}^k_i(\bm{x}, \bm{P}) \big )} .
\end{equation}
As illustrated in Fig.~\ref{fig:net}(b), our method not only matches the predicted gradient on the position of $\bm{x}$, but also matches the gradient on the neighboring regions of $\bm{x}$.
This is important because our input point cloud is noisy and individual points may not lie on the underlying surface.
Finally, the training loss is an aggregation of the objective for each neural gradient learning function $\mathcal{L}(\bm{\theta})|_{\bm{x}_i}$ of $\bm{x}_i$, \ie,
\begin{equation}
   \mathcal{L}_{\text{NGL}} = \frac{1}{M_1}\sum_{i=1}^{M_1} \mathcal{L}(\bm{\theta})|_{\bm{x}_i} ~, ~\bm{x}_i \in \bm{X}.
\end{equation}

For the distribution $\mathcal{D}$, we make it concentrate in the neighborhood of $\bm{x}$ in 3D space.
Specifically, $\mathcal{D}$ is set by uniform sampling points $\bm{x}$ from $\bm{P}$ and placing an isotropic Gaussian $N(\bm{x},\sigma^2)$ for each $\bm{x}$.
The distribution parameter $\sigma$ depends on each point $\bm{x}$ and is adaptively set to the distance from the $50$th nearest point to $\bm{x}$~\cite{atzmon2020sal,atzmon2021sald}.

\begin{figure}[t]
   \centering
   \includegraphics[width=\linewidth]{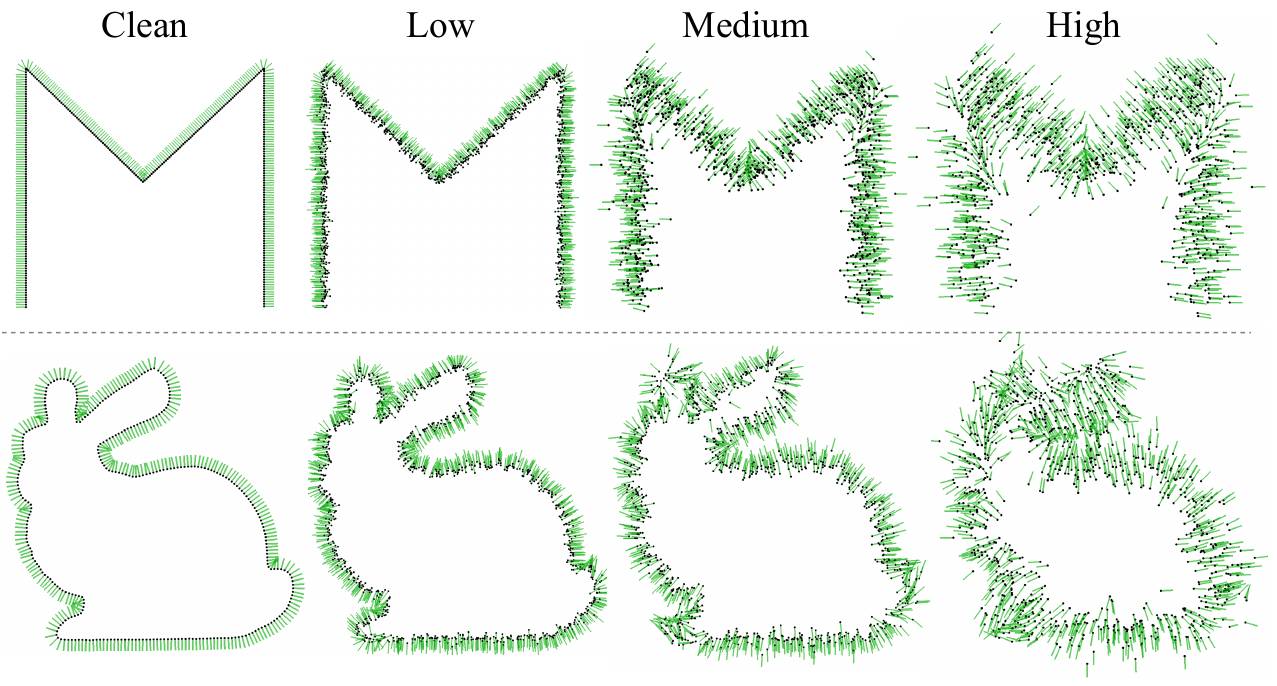}  \vspace{-0.75cm}
   \caption{
      Our method can estimate gradient vectors (green rays) from point clouds (black dots) with different noise levels.
   }
   \label{fig:grad}
   \vspace{-0.2cm}
\end{figure}

Our network architecture for neural gradient learning is based on the one used in~\cite{atzmon2020sal,ma2020neural}, which is composed of eight linear layers with ReLU activation functions (except the last layer) and a skip connection.
After training, the network can derive pointwise gradients from the raw data $\bm{P}$ (see 2D examples in Fig.~\ref{fig:grad}).

\noindent\textbf{Extension}.
If we assume the raw data $\bm{P}$ is noise-free, that is, the neighbors $\mathcal{N}^k(\bm{x}, \bm{P})$ are located on the surface, then the formula of Eq.~\eqref{eq:loss_grad} can take another form
\begin{equation} \label{eq:loss_offset}
   \mathcal{L}(\bm{\theta}) = \norm{\big (f(\bm{x}; \bm{\theta}) \cdot \bm{v} - \bm{x} \big ) + \frac{1}{k}\sum_{i=1}^{k} \mathcal{N}^k_i(\bm{x}, \bm{P})} .
\end{equation}
More particularly, if we set $k\!=\!1$ and the nearest point of $\bm{x}$ in $\bm{P}$ be $\bm{p}$, \ie, $\mathcal{N}^{k=1}(\bm{x}, \bm{P}) \!=\! \bm{p}$, then the above formula is turned into the special case in Eq.~\eqref{eq:pull}.
Specifically, the derived formula in Eq.~\eqref{eq:loss_offset} also distinguishes our method from the methods~\cite{ma2020neural,chibane2020neural,atzmon2020sal,atzmon2021sald}, since their objectives only consider the location of each clean point, while our proposed objective covers the neighborhood of each noisy point to approximate the surface gradients.

%%% -------------------------------------------------------------------------
\subsection{Gradient Vector Optimization} \label{sec:optim}

\begin{figure}[t]
   \centering
   \includegraphics[width=0.9\linewidth]{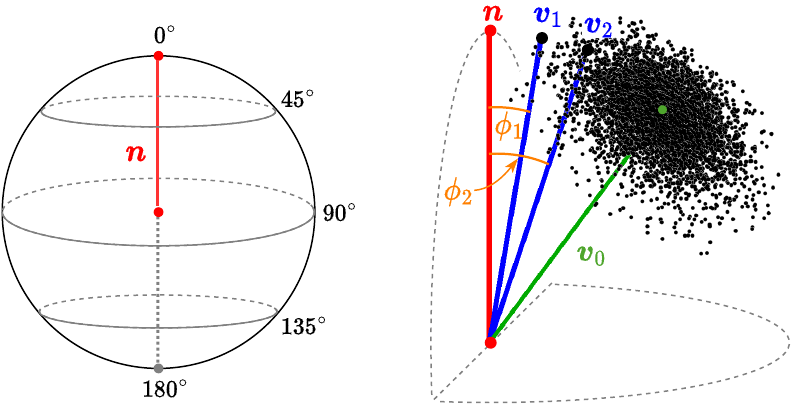}  \vspace{-0.4cm}
   \caption{
      \emph{Left}: illustration of the angular distance field of a vector {\color{red}$\bm{n}$}.
      \emph{Right}: given an initial vector {\color[RGB]{79,166,45}$\bm{v}_0$} and its vector samples in the unit sphere (black dots with a Gaussian distribution), our method will select vector {\color{blue}$\bm{v}_1$} rather than {\color{blue}$\bm{v}_2$} as a candidate since {\color{blue}$\bm{v}_1$} has a smaller angular distance {\color[RGB]{255,128,0}$\phi$} with respect to the target vector {\color{red}$\bm{n}$}.
   }
   \label{fig:optim}
   \vspace{-0.2cm}
\end{figure}

A notable shortcoming of neural gradient learning is that the derived gradient vectors are inaccurate because the implicit function tries to approximate the whole shape surface instead of focusing on fitting local regions.
Therefore, the learned gradient vectors are inadequate to be used as surface normals and need to be further refined.
Inspired by the implicit surface representations, we define the expected normal as the zero level-set of a function
\begin{equation}
   \bm{\mathcal{V}} = \left \{\bm{x} \in \mathbb{R}^3, \bm{v} \in \mathbb{R}^3 ~ \vert ~ g(\bm{x},\bm{v}; \bm{\vartheta}) = 0 \right \},
\end{equation}
where $g\colon\mathbb{R}^{3}\times \mathbb{R}^3 \rightarrow \mathbb{R}$ is a neural network with parameter $\bm{\vartheta}$ that predicts (unsigned) angular distance field between the normalized gradient vector $\bm{v}$ and the ground-truth normal vector $\bm{\hat{n}}$ (see Fig.~\ref{fig:optim}).
Given appropriate training objectives, the zero level-set of $g$ can be a vector cluster describing the normals of point cloud $\bm{P}$.
To this end, we introduce \emph{Gradient Vector Optimization} (GVO) defined by the form of a loss function
\begin{equation}
   \mathcal{L}(\bm{\vartheta}) = \mathbb{E}_{\bm{v} \sim \mathcal{D'}} ~\mathcal{T} \big(g(\bm{x},\bm{v}; \bm{\vartheta}),~ \langle\bm{v}, \bm{\hat{n}}\rangle \big),
\end{equation}
where $\mathcal{D'}$ is a probability distribution based on an initial vector $\bm{v} \!\in\! \mathbb{R}^3$.
$\langle\cdot\rangle \!\in\! [0,\pi]$ means the angular difference between two unit vectors.
In contrast to the previous method~\cite{li2023NeAF}, we regress angles using weighted features of the approximated local plane instead of point features from PointNet~\cite{qi2017pointnet}.
The motivation is that simple angle regression with $g$ fails to be robust to noise or produce high-quality normals.

Given a neighborhood size $m$, we can construct the input data as the nearest neighbor graph $G \!=\! (\mathcal{N}, \mathcal{E})$, where $(\bm{x},\bm{x}_j) \!\in\! \mathcal{E}$ is a directed edge if $\bm{x}_j$ is one of the $m$ nearest neighbors of $\bm{x}$.
Let $\mathcal{N}^{m}(\bm{x}) \!=\! \{\bm{x}_j - \bm{x}\}_{j=1}^m$ be the centered coordinates of the points in the neighborhood.
The standard way to solve for unoriented normal at a point is to fit a plane to its local neighborhood~\cite{levin1998approximation}, which is described as
\begin{equation} \label{eq:plane}
   \bm{n}_i^\ast = \mathop{\text{argmin}}_{\bm{n}} \sum_{\bm{x}_j'\in\mathcal{N}^{m}(\bm{x}_i)} \norm{\bm{x}_j' \cdot \bm{n}}^2.
\end{equation}
In practice, there are two main issues about the utilizing of Eq.~\eqref{eq:plane} \cite{lenssen2020deep}:
(\romannumeral1) it acts as a low-pass filter for the data and eliminates sharp details,
(\romannumeral2) it is unreliable if there is noise or outliers in the data.
We will show that both issues can be resolved by integrating weighting functions into our optimization pipeline.
In short, the preservation of detailed features is achieved by an anisotropic kernel that infers weights of point pairs based on their relative positions, while the robustness to outliers is achieved by a scoring mechanism that weights points according to inlier scores.

\begin{figure}[t]
   \centering
   \includegraphics[width=\linewidth]{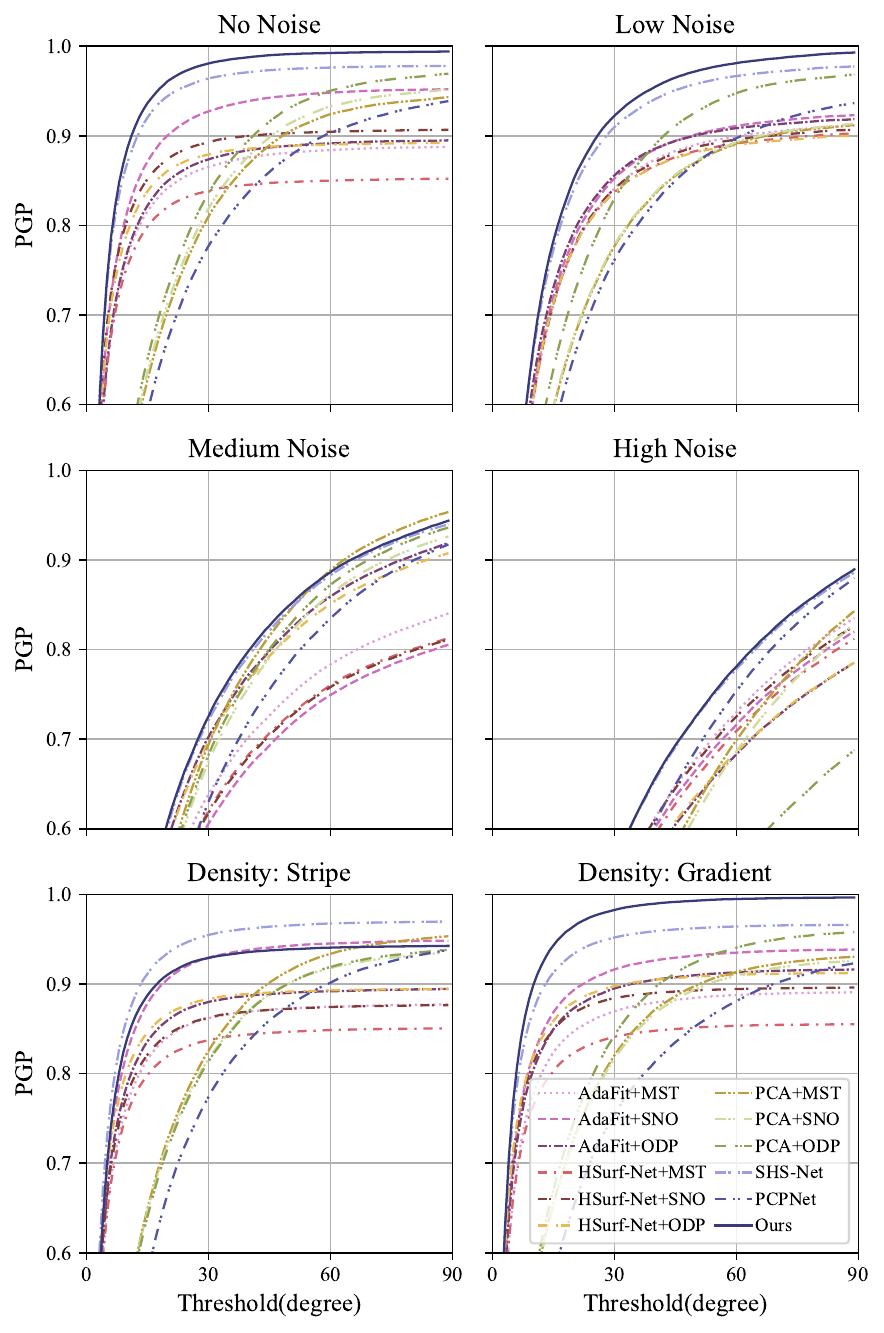}  \vspace{-0.6cm}
   \caption{
    The PGP curves of oriented normal on the FamousShape dataset.
    It depicts the percentage of good points (PGP) for a given angle threshold.
    Our method achieves the best value at most of the thresholds.
   }
   \label{fig:curve_FamousShape}
   \vspace{-0.2cm}
\end{figure}

\input{table_pcpnet_famousShape_orient.tex}
\input{table_pcpnet_famousShape_unorient.tex}
\input{table_time.tex}

\noindent\textbf{Anisotropic Kernel}.
For feature encoding, our extraction layer is formulated as
\begin{equation}
   \bm{x}_{l}' \!=\! \gamma\left(\bm{x}_{l},~ \beta\left(\text{MAX} \left\{\alpha(w_{j} \cdot \bm{x}_{j}) \right\}_{j=1}^{m} \right)\right),
   ~ l \!=\! 1,\cdots,m',
\end{equation}
where $\text{MAX}\{\cdot\}$ indicates the feature maxpooling over the neighbors $\mathcal{N}^m(\bm{x}) \!=\! \{\bm{x_j}-\bm{x}\}_{j=1}^m$ of a center point $\bm{x}$.
$m' \!\leqslant\! m$ means that fewer neighbors are used in the next layer, and we usually set $m'$ to $m/2$.
$\alpha, \beta$ and $\gamma$ are MLPs.
They compose an anisotropic kernel that considers the full geometric relationship between neighboring points, not just their positions, thus providing features with richer contextual information.
Specifically, $w$ is a weight given by
\begin{equation}  \label{eq:weight}
   w_{j} = \frac{d_j}{\sum_{i=1}^{m}{d_i}}, ~ d_i = \text{sigmoid}\big(\vartheta_1 - \vartheta_2{\norm{\bm{x}_i - \bm{x}}} \big),
\end{equation}
where $\vartheta_1$ and $\vartheta_2$ are learnable parameters with the initial value set to 1.
The weight $w$ lets the kernel concentrate on the points $\bm{x}_i \in \mathcal{N}^m(\bm{x})$ that are closer to its center $\bm{x}$.

\noindent\textbf{Inlier Score}.
Based on the neighbors $\mathcal{N}^m(\bm{x})$ of $\bm{x}$, the inlier score function $s(\bm{x},\bm{v}; \bm{\vartheta})$ is optimized by
\begin{equation}
   \mathcal{L}_1(\bm{\vartheta}) = \mathbb{E}_{\bm{v} \sim \mathcal{D'}} ~ \mathcal{T}_1 \big(s(\bm{x}_i,\bm{v}; \bm{\vartheta}),~ \delta(\bm{x}_i, \bm{\hat{n}}) \big),
   ~ \bm{x}_i \in \mathcal{N}^m(\bm{x}) ~,
\end{equation}
where $\mathcal{T}_1$ is mean squared error.
The function $s$ assigns low scores to outliers and high scores to inliers.
Correspondingly, $\delta$ generates scores based on the distance between neighboring points $\bm{x}_i$ and the local plane determined by the normal vector $\bm{\hat{n}}$ at point $\bm{x}$, that is
\begin{equation}
   \delta(\bm{x}_i, \bm{\hat{n}}) = \text{exp} \left(- \frac{(\bm{x}_i \cdot \hat{\bm{n}})^2}{\rho^2} \right),~ \bm{x}_i \in \mathcal{N}^m(\bm{x})~,
\end{equation}
where $\rho \!=\! \text{max} (0.05^2, ~0.3 \sum_{i=1}^{m}(\bm{x}_i \cdot \hat{\bm{n}})^2 / m )$~\cite{li2022hsurf}.
The function $s$ regresses the score of each point in the neighbor graph, and these scores are used to find the vector angles based on score-weighted gradient vector optimization
\begin{equation} \label{eq:grad_opti}
   \mathcal{L}_2(\bm{\vartheta}) = \mathbb{E}_{\bm{v} \sim \mathcal{D'}} ~\mathcal{T}_2 \big(s \odot g(\bm{x},\bm{v}; \bm{\vartheta}),~ \langle\bm{v}, \bm{\hat{n}}\rangle \big),
\end{equation}
where $\mathcal{T}_2$ is mean absolute error.
$\odot$ denotes that the score function $s$ is integrated into the feature encoding of learning angular distance field.
The score and angle are jointly regressed by MLP layers based on the neighbor graph.
In summary, our final training loss is
\begin{equation} \label{eq:loss_optim}
  \mathcal{L}_{\text{GVO}} = \mathcal{L}_{1}(\bm{\vartheta}) + \lambda \mathcal{L}_{2}(\bm{\vartheta}) ~,
\end{equation}
where $\lambda\!=\!0.5$ is a weighting factor.

\noindent\textbf{Distribution $\mathcal{D'}$}.
This distribution is different during the training and testing phases.
During training, we first uniformly sample $M_2$ random vectors in 3D space for each point of the input point cloud.
Then the network is trained to predict the angle of each vector with respect to the ground-truth normal.
At test time, we establish an isotropic Gaussian $N\big(\bm{v}, (\eta\cdot45^\circ)^2 \big)$ that forms a distribution about the initial gradient vector $\bm{v}$ in the unit sphere, and then we obtain a set of $M_3$ vector samples around $\bm{v}$.
As shown in Fig.~\ref{fig:optim}, the trained network tries to find an optimal candidate as output from the vector samples according to the predicted angle.

%%% =================================================================================================
\section{Experiments}

\noindent\textbf{Implementation}.
For NGL, the $k$ in Eq.~\eqref{eq:ave_grad} is set to $k \!=\! 64$ and we select $M_1 \!=\! 5000$ points from distribution $\mathcal{D}$ as the input during training.
For GVO, we train it only on the PCPNet training set~\cite{guerrero2018pcpnet} and use the provided normals to calculate vector angles.
We select $m \!=\! 700$ neighboring points for each query point.
For the distribution $\mathcal{D'}$, we set $M_2 \!=\! 500$, $M_3 \!=\! 4000$ and $\eta \!=\! 0.4$.

\noindent\textbf{Metrics}.
We use the Root Mean Squared Error (RMSE) to evaluate the estimated normals and use the Percentage of Good Points (PGP) to show the error distribution~\cite{zhu2021adafit,li2022hsurf}.
% which is plotted by the Percentage of Good Point normals (PGP) with errors below different angle thresholds.
% Note that for the baseline methods, we flip their oriented normals if more than half of the normals face inward.

\input{table_ablation.tex}

\begin{figure}[t]
   \centering
   \vspace{-0.15cm}
   \includegraphics[width=\linewidth]{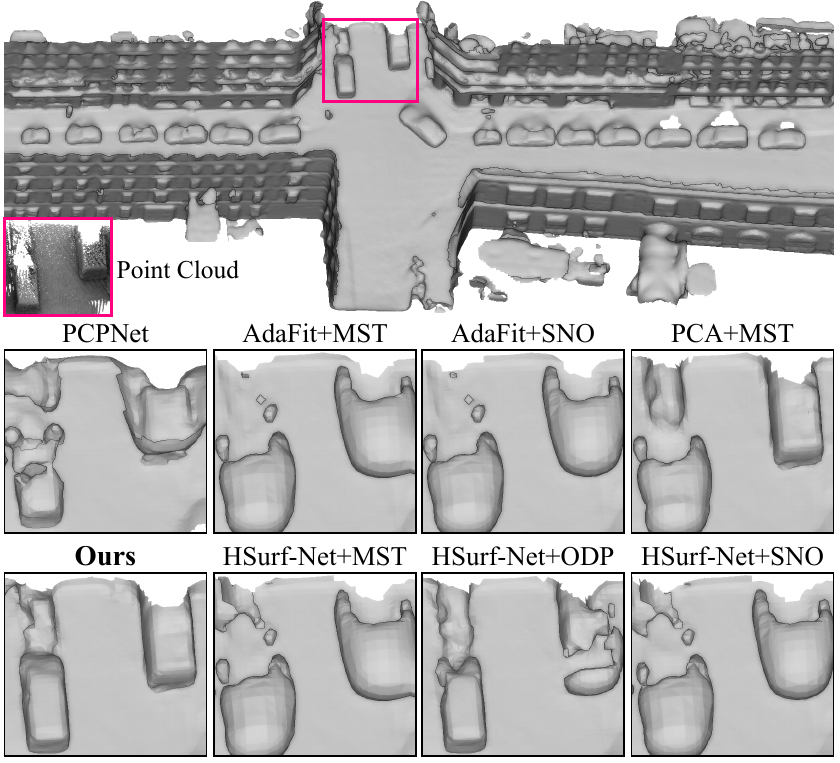}  \vspace{-0.6cm}
   \caption{
      The top row shows the scene reconstructed from LiDAR data using our estimated normals, and below is a local region comparison of the different methods.
   }
   \label{fig:poisson}
   \vspace{-0.2cm}
\end{figure}

\begin{figure}[t]
   \centering
   \includegraphics[width=\linewidth]{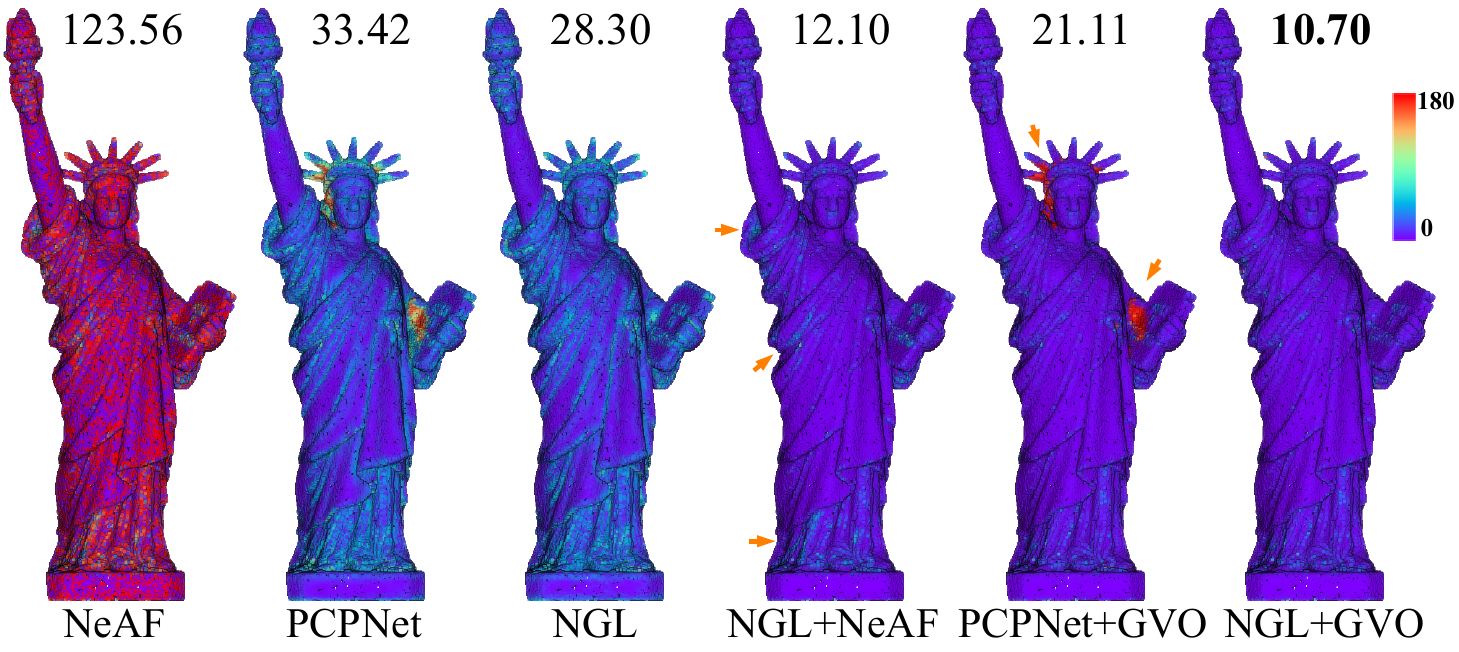}  \vspace{-0.7cm}
   \caption{
      Error maps of oriented normals.
      We integrate our NGL and GVO into other methods to estimate oriented normals.
      The mean value of RMSE is provided above each shape.
   }
   \label{fig:errorMap_2stages}
   \vspace{-0.2cm}
\end{figure}

%-------------------------------------------------------------------------
\subsection{Evaluation}

\noindent\textbf{Evaluation of Oriented Normal}.
The baseline methods include PCPNet~\cite{guerrero2018pcpnet}, DPGO~\cite{wang2022deep}, SHS-Net~\cite{li2023shsnet} and different two-stage pipelines, which are built by combining unoriented normal estimation methods (PCA \cite{hoppe1992surface}, AdaFit~\cite{zhu2021adafit}, HSurf-Net~\cite{li2022hsurf}) and normal orientation methods (MST~\cite{hoppe1992surface}, SNO~\cite{schertler2017towards}, ODP~\cite{metzer2021orienting}).
We choose them as they are representative algorithms in this research field at present.
The quantitative comparison results on datasets PCPNet~\cite{guerrero2018pcpnet} and FamousShape~\cite{li2023shsnet} are shown in Table~\ref{table:pcpnet_famousShape_o}.
It is clear that our method achieves large performance improvements over the vast majority of noise levels and density variations on both datasets.
Through this experiment, we also find that combining a better unoriented normal estimation algorithm with the same normal orientation algorithm does not necessarily lead to better orientation results, \eg, PCA+MST \textit{vs}. AdaFit+MST and PCA+SNO \textit{vs}. HSurf-Net+SNO.
The error distributions in Fig.~\ref{fig:curve_FamousShape} show that our method has the best performance at most of the angle thresholds.

We provide more experimental results on different datasets in the supplementary material, including comparisons with GCNO~\cite{xu2023globally} on sparse data and more applications to surface reconstruction.

\noindent\textbf{Evaluation of Unoriented Normal}.
In this evaluation, we ignore the orientation of normals and compare our method with baselines that are used for estimating unoriented normals, such as the traditional methods PCA~\cite{hoppe1992surface} and Jet~\cite{cazals2005estimating}, the learning-based surface fitting methods AdaFit~\cite{zhu2021adafit} and GraphFit~\cite{li2022graphfit}, and the learning-based regression methods NeAF~\cite{li2023NeAF} and HSurf-Net~\cite{li2022hsurf}.
The quantitative comparison results on datasets PCPNet~\cite{guerrero2018pcpnet} and FamousShape~\cite{li2023shsnet} are reported in Table~\ref{table:pcpnet_famousShape}.
We can see that our method has the best performance under most point cloud categories and achieves the best average result.

\noindent\textbf{Application}.
We employ the Poisson reconstruction algorithm \cite{kazhdan2013screened} to generate surfaces from the estimated oriented normals on the Paris-rue-Madame dataset~\cite{serna2014paris}, acquired from the real-world using laser scanners.
The reconstructed surfaces are shown in Fig.~\ref{fig:poisson}, where ours exhibits more complete and clear car shapes.

\noindent\textbf{Complexity and Efficiency}.
We evaluate the learning-based oriented normal estimation methods on a machine equipped with NVIDIA 2080 Ti GPU.
In Table~\ref{tab:time}, we report the RMSE, number of learnable network parameters, and test runtime for each method on the PCPNet dataset.
Our method achieves significant performance improvement with minimal parameters and relatively less runtime.

%%% -------------------------------------------------------------------------
\subsection{Ablation Studies}  \label{sec:ablation}
Our method seeks to achieve better performance in both unoriented and oriented normal estimation.
We provide the ablation results of our method in Table~\ref{tab:ablation} (a)-(e), which are discussed as follows.

\noindent\textbf{(a) Component}.
We remove NGL, GVO, inlier score and weight $w$ of the anisotropic kernel, respectively.
If NGL is not used, we optimize a randomly sampled set of vectors in the unit sphere for each point, but the optimized normal vectors face both sides of the surface, resulting in the worst orientations.
Gradient vectors from NGL are inaccurate when used as normals without being optimized by GVO.
The score and weight are important for improving performance, especially in unoriented normal evaluation.

\noindent\textbf{(b) Loss}.
Replacing L2 distance in $\mathcal{L}_{\sss \text{NGL}}$ with L1 distance or MSE is not a good choice.
We also alternatively set $\lambda$ in $\mathcal{L}_{\sss \text{GVO}}$ to $0.2$ or $0.8$, both of which lead to worse results.

\noindent\textbf{(c) Size $k$}.
For the neighborhood size in Eq.~\eqref{eq:ave_grad}, we alternatively set $k$ to $1$, $32$ or $128$, however, all of which do not bring better oriented normal results.

\noindent\textbf{(d) Distribution $\mathcal{D}$}.
We change the distribution parameter $\sigma$ as the distance $d_{\sigma}$ of the $32$th or $64$th nearest point to $\bm{x}$, whereas the results get worse.

\noindent\textbf{(e) Distribution $\mathcal{D'}$}.
We change the distribution parameter $\eta$ to $0.3$ or $0.5$ and the vector sample size $M_2$ to $3000$ or $5000$, respectively.
The influence of these parameters on the results is relatively small.
The larger size gives better results, but requires more time and memory consumption.

\noindent\textbf{(f) Modularity}.
In Fig.~\ref{fig:errorMap_2stages}, we show that our NGL and GVO can be integrated into some other methods (PCPNet~\cite{guerrero2018pcpnet} and NeAF~\cite{li2023NeAF}) to estimate more accurate oriented normals.
Note that NeAF can not estimate oriented normals.
We can see that our NGL+GVO gives the best results.

%%% =================================================================================================
\section{Conclusion}

In this work, we propose to learn neural gradient from point cloud for oriented normal estimation.
We introduce \emph{Neural Gradient Learning} (NGL) and \emph{Gradient Vector Optimization} (GVO), defined by a family of loss functions.
Specifically, we minimize the corresponding loss to let the NGL learn gradient vectors from global geometry representation, and the GVO optimizes vectors based on an insight into the local property.
Moreover, we integrate two weighting functions, including anisotropic kernel and inlier score, into the optimization to improve robust and detail-preserving performance.
We provide extensive evaluation and ablation experiments that demonstrate the state-of-the-art performance of our method and the effectiveness of our designs.
Future work includes improving the performance under high noise and density variation, and exploring more application scenarios of our algorithm.

% DO NOT INCLUDE ACKNOWLEDGMENTS IN AN ANONYMOUS SUBMISSION TO SIGGRAPH 2019
\begin{acks}

This work was supported by National Key R\&D Program of China (2022YFC3800600), the National Natural Science Foundation of China (62272263, 62072268), and in part by Tsinghua-Kuaishou Institute of Future Media Data.

\end{acks}

%%% Bibliography
\bibliographystyle{ACM-Reference-Format}
\bibliography{egbib}



\section*{Supplementary material}

# Neural Gradient Learning and Optimization for Oriented Point Normal Estimation —— *Supplementary Material*

**Qing Li**
School of Software, Tsinghua University
Beijing, China
leoqli@tsinghua.edu.cn

**Huifang Feng**
School of Informatics, Xiamen University
Xiamen, China
fenghuifang@stu.xmu.edu.cn

**Kanle Shi**
Kuaishou Technology
Beijing, China
shikanle@kuaishou.com

**Yi Fang**
Center for Artificial Intelligence and Robotics, New York University
Abu Dhabi, UAE
yfang@nyu.edu

**Yu-Shen Liu**[*]
School of Software, Tsinghua University
Beijing, China
liuyushen@tsinghua.edu.cn

**Zhizhong Han**
Department of Computer Science, Wayne State University
Detroit, USA
h312h@wayne.edu

## 1 EVALUATION METRICS

As in [Ben-Shabat and Gould 2020; Guerrero et al. 2018; Li et al. 2022; Zhu et al. 2021], in order to evaluate the estimated normal vectors, we use the Root Mean Squared Error (RMSE) of vector angles between the ground-truth normals $\hat{\mathbf{n}}_i$ and the estimated normals $\mathbf{n}_i$, *i.e.*,

$$\text{RMSE}_U = \sqrt{\frac{1}{M}\sum_{i=1}^{M}\left(\arccos(|\hat{\mathbf{n}}_i \odot \mathbf{n}_i|)\right)^2}, \qquad (1)$$

$$\text{RMSE}_O = \sqrt{\frac{1}{M}\sum_{i=1}^{M}\left(\arccos(\hat{\mathbf{n}}_i \odot \mathbf{n}_i)\right)^2}, \qquad (2)$$

where $\text{RMSE}_U$ and $\text{RMSE}_O$ are evaluation metrics for unoriented and oriented normals, respectively. $M$ is the number of evaluated point normals. $|\cdot|$ means the absolute value of the inner product $\odot$ of two normal vectors. The range of normal angle errors is bounded between $0°$ and $90°$ in unoriented normal evaluation, and between $0°$ and $180°$ in oriented normal evaluation.

Furthermore, we also employ the Percentage of Good Points (PGP) to analyze the distribution of normal errors. It depicts the percentage of good points whose normal angle errors are less than the given angle thresholds. Specifically, the PGP is calculated by

$$\text{PGP}_U(\tau) = \frac{1}{M}\sum_{i=1}^{M}\mathcal{J}\left(\arccos(|\hat{\mathbf{n}}_i \odot \mathbf{n}_i|) < \tau\right), \qquad (3)$$

$$\text{PGP}_O(\tau) = \frac{1}{M}\sum_{i=1}^{M}\mathcal{J}\left(\arccos(\hat{\mathbf{n}}_i \odot \mathbf{n}_i) < \tau\right), \qquad (4)$$

where $\text{PGP}_U(\tau)$ and $\text{PGP}_O(\tau)$ are metrics for unoriented and oriented normal evaluation, respectively. $\mathcal{J}$ denotes an indicator function used to measure whether the angle error is less than a given threshold $\tau$.

## 2 MORE ANALYSIS ON NGL AND GVO

### 2.1 NGL

**Ablations about $k$.** We provide more ablation results about the parameter $k$ in our NGL. As shown in Table 1 and Fig. 1, we report

---
*The corresponding author is Yu-Shen Liu.

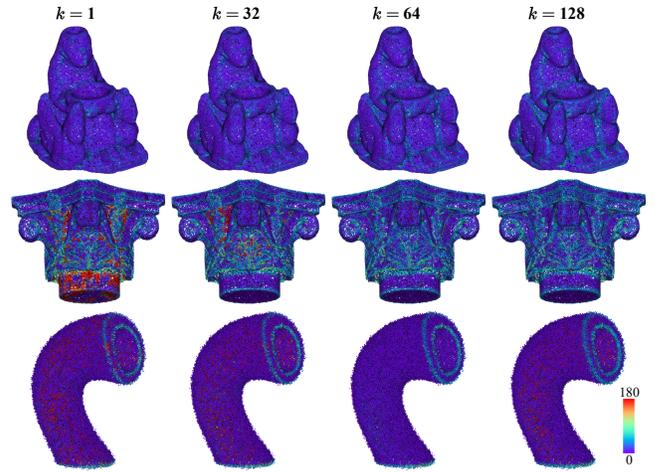

**Figure 1: Error visualization of gradients from NGL with different $k$. For noise-free point clouds, smaller values of $k$ can provide more accurate normals, but may bring wrong orientations in some noisy cases.**

quantitative and qualitative results of the learned gradients (initial oriented normals) by NGL. In Table 1, we can see that smaller values of $k$ give better unoriented normals, and these better results mainly come from the noise-free point clouds, such as the categories of none noise, stripe density and gradient density. For oriented normal estimation, a relatively large value can lead to better results, and $k = 64$ gives the best oriented normal results. A large value is more robust to high noise (such as 0.6% and 1.2%) and improves the performance, but a too large value, *e.g.*, $k = 128$, degrades the performance. Furthermore, we can observe from Table 1 that the advantage of smaller $k$ values in unoriented normal estimation vanishes after optimizing the initial normals using our GVO. For NGL+GVO, values 1, 32 and 64 of $k$ all have the same average RMSE for unoriented normals, but $k = 64$ still has the best oriented normal results. Thus, we set $k$ to 64 in our algorithm. In Fig. 2, we show the oriented normal errors of NGL on different categories of shapes. It shows that our NGL can provide oriented normals with consistent orientations on various point cloud data.

**Table 1: Ablation studies of $k$ in our NGL under the metric of unoriented and oriented normal on the PCPNet dataset.**

| Category | | | Unoriented Normal | | | | | | | | Oriented Normal | | | | | |
|---|---|---|---|---|---|---|---|---|---|---|---|---|---|---|---|---|
| | | | Noise | | | Density | | Average | | | Noise | | | Density | | Average |
| | | None | 0.12% | 0.6% | 1.2% | Stripe | Gradient | | None | 0.12% | 0.6% | 1.2% | Stripe | Gradient | |
| NGL | $k=1$ | 9.51 | 11.49 | 21.79 | 27.53 | 10.25 | 9.80 | **15.06** | 16.51 | 19.05 | 38.31 | 46.53 | 13.09 | 12.81 | 24.38 |
| | $k=32$ | 11.18 | 11.87 | 18.35 | 25.50 | 13.15 | 12.10 | 15.36 | 18.31 | 25.50 | 29.42 | 34.32 | 22.48 | 14.05 | 24.01 |
| | $k=64$ | 12.24 | 12.74 | 17.89 | 23.88 | 15.16 | 13.75 | 15.94 | 18.39 | 15.32 | 25.20 | 32.57 | 22.91 | 15.73 | **21.69** |
| | $k=128$ | 13.98 | 14.44 | 18.09 | 22.99 | 20.04 | 16.23 | 17.63 | 20.06 | 27.12 | 34.44 | 31.91 | 40.74 | 25.02 | 29.88 |
| NGL +GVO | $k=1$ | 4.07 | 8.70 | 16.13 | 21.65 | 4.79 | 4.55 | **9.98** | 13.57 | 18.24 | 38.29 | 47.23 | 9.27 | 8.99 | 22.60 |
| | $k=32$ | 4.06 | 8.69 | 16.13 | 21.65 | 4.79 | 4.56 | **9.98** | 13.64 | 24.31 | 29.83 | 33.93 | 17.37 | 8.51 | 21.27 |
| | $k=64$ | 4.06 | 8.70 | 16.12 | 21.65 | 4.80 | 4.56 | **9.98** | 12.52 | 12.97 | 25.94 | 33.25 | 16.81 | 9.47 | **18.49** |
| | $k=128$ | 4.08 | 8.70 | 16.13 | 21.64 | 4.84 | 4.58 | 9.99 | 12.84 | 23.65 | 34.96 | 33.03 | 37.64 | 18.42 | 26.76 |

**Table 2: Comparison of the learned gradients under the metric of unoriented and oriented normal on the PCPNet dataset**

| Category | | | Unoriented Normal | | | | | | | | Oriented Normal | | | | | |
|---|---|---|---|---|---|---|---|---|---|---|---|---|---|---|---|---|
| | | | Noise | | | Density | | Average | | | Noise | | | Density | | Average |
| | | None | 0.12% | 0.6% | 1.2% | Stripe | Gradient | | None | 0.12% | 0.6% | 1.2% | Stripe | Gradient | |
| NGL | $k=1$ | 9.51 | 11.49 | 21.79 | 27.53 | 10.25 | 9.80 | **15.06** | 16.51 | 19.05 | 38.31 | 46.53 | 13.09 | 12.81 | 24.38 |
| | $k=64$ | 12.24 | 12.74 | 17.89 | 23.88 | 15.16 | 13.75 | 15.94 | 18.39 | 15.32 | 25.20 | 32.57 | 22.91 | 15.73 | **21.69** |
| NP [Ma et al. 2021] | - | 9.29 | 13.75 | 21.47 | 26.17 | 10.77 | 9.87 | 15.22 | 16.54 | 28.84 | 33.69 | 38.95 | 18.42 | 12.72 | 24.86 |

**Table 3: We integrate our NGL and GVO into other methods (PCPNet [Guerrero et al. 2018] and NeAF [Li et al. 2023b]) to estimate oriented normals. Note that NeAF can not estimate oriented normals.**

| | Category | | Unoriented Normal | | | | | | | Oriented Normal | | | | | | |
|---|---|---|---|---|---|---|---|---|---|---|---|---|---|---|---|---|
| | | None | Noise | | | Density | | Average | None | Noise | | | Density | | Average |
| | | | 0.12% | 0.6% | 1.2% | Stripe | Gradient | | | 0.12% | 0.6% | 1.2% | Stripe | Gradient | |
| **(a)** | NeAF | 4.20 | 9.25 | 16.35 | 21.74 | 4.89 | 4.88 | 10.22 | 125.21 | 122.71 | 119.69 | 117.22 | 125.01 | 124.25 | 122.35 |
| | PCPNet | 14.07 | 15.01 | 21.14 | 25.59 | 16.42 | 17.27 | 18.25 | 33.34 | 34.22 | 40.54 | 44.46 | 37.95 | 35.44 | 37.66 |
| | NGL | 12.24 | 12.74 | 17.89 | 23.88 | 15.16 | 13.75 | 15.94 | 18.39 | 15.32 | 25.20 | 32.57 | 22.91 | 15.73 | 21.69 |
| **(b)** | NGL+NeAF | 4.26 | 9.27 | 16.35 | 21.73 | 4.95 | 4.93 | 10.25 | 12.59 | 13.57 | 26.10 | 33.30 | 16.88 | 9.26 | 18.62 |
| | PCPNet+GVO | 4.08 | 8.71 | 16.13 | 21.65 | 4.81 | 4.61 | 10.00 | 30.42 | 33.18 | 41.34 | 45.34 | 35.27 | 32.16 | 36.29 |
| | NGL+GVO | 4.06 | 8.70 | 16.12 | 21.65 | 4.80 | 4.56 | **9.98** | 12.52 | 12.97 | 25.94 | 33.25 | 16.81 | 9.47 | **18.49** |

**Table 4: Comparison of oriented normal estimation on sparse point clouds. The algorithms of GCNO and PCA+MST run on the CPU, and the test runtime (seconds per 5000 points) of GCNO is much longer than other methods. Our method has the lowest RMSE and relatively high efficiency.**

| | PCA+MST [Hoppe et al. 1992] | HSurf-Net [Li et al. 2022] +ODP [Metzer et al. 2021] | PCPNet [Guerrero et al. 2018] | GCNO [Xu et al. 2023] | SHS-Net [Li et al. 2023a] | Ours |
|---|---|---|---|---|---|---|
| RMSE | 45.40 | 62.51 | 48.48 | 41.24 | 32.64 | **28.34** |
| Time | **0.01+0.71** | 3.87+31.75 | 3.34 | 822.60 | 3.54 | 2.53 |

**Comparison of gradients**. As we describe in our paper, in implicit function learning, the gradient is the direction in which the distance value increases the fastest. Some surface reconstruction methods exploit this property to move a query position $x$ by distance $f(x; \theta)$ along or against the gradient direction $v$ to its closest point $p$ sampled from the manifold. Specifically, $f(x; \theta)$ is interpreted as a signed distance in NP [Ma et al. 2021], which can learn highly accurate signed distance functions directly from the input noise-free point clouds. Here we compare the learned gradients in NP with our method. As shown in Table 2 and Fig. 3, we provide quantitative and qualitative comparison results of our NGL and NP, respectively. Our NGL can deliver more accurate results when directly using the learned gradients as surface normals. Our method is good at learning a neural gradient field with a consistent orientation from a variety of point cloud data, even in the presence of noise. In contrast, NP tries to learn an accurate distance field to approximate the underlying surface of point cloud data.

## 2.2 GVO

In our algorithm, the learned gradients from NGL are further optimized by GVO to predict oriented normals. From the quantitative and qualitative results shown in Table 1 and Fig. 4, we can easily conclude that GVO largely improves the accuracy of the oriented normals based on the learned gradients from NGL.

In our experiments, we use the model trained in 800 epochs. Here we evaluate our GVO on the PCPNet test set using models trained for 100 to 1000 epochs. The estimated normals are measured using the evaluation metrics $\text{RMSE}_U$ and $\text{RMSE}_O$. As shown in Fig. 5, we plot the average evaluation results at different noise levels and different density variations. It can be seen from the

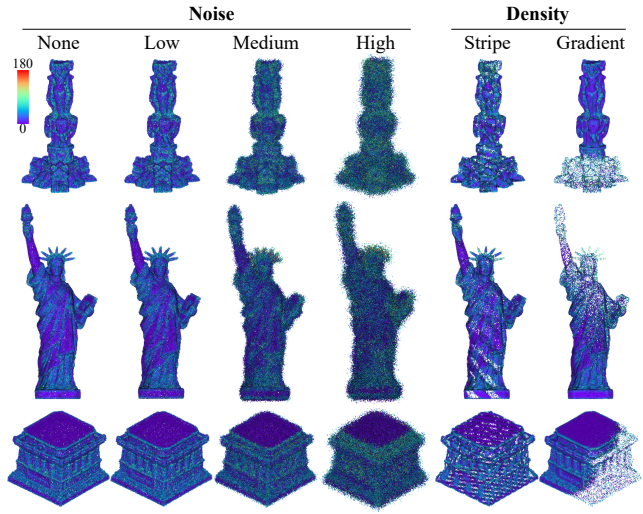

**Figure 2: Error maps of gradients (*i.e.*, the coarse normals) from our NGL on different categories of shapes. Our NGL can provide gradients with consistent orientations.**

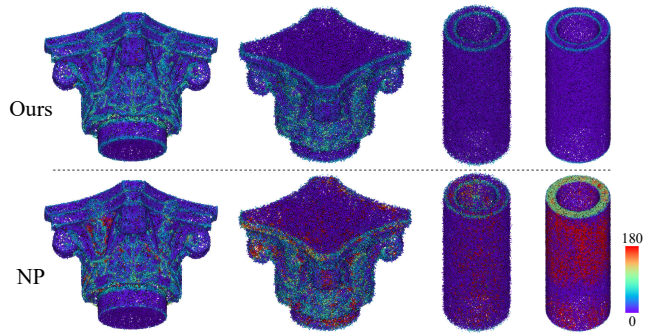

**Figure 3: Comparison of the learned gradients from our NGL and NP [Ma et al. 2021]. Our method has clear advantages over NP, especially in the presence of noise.**

curves in Fig. 5 that the errors in unoriented normal evaluation keep decreasing, while there are some fluctuations in the errors of oriented normal evaluation. After about 800 epochs of training, the errors of both unoriented and oriented normal evaluations have no obvious change.

### 2.3 Modularity of NGL and GVO

As we describe in the paper, our NGL and GVO can be integrated into some other methods (such as PCPNet [Guerrero et al. 2018] and NeAF [Li et al. 2023b]) to estimate more accurate oriented normals. In Table 3 and Fig. 6, we show more quantitative and qualitative results to further demonstrate this advantage of our approach. We can see that our NGL lets the NeAF have the capability of estimating oriented normals, and our GVO largely improves the accuracy of unoriented normals from PCPNet. In summary, our NGL+GVO leads to the best results for both unoriented and oriented normal estimations.

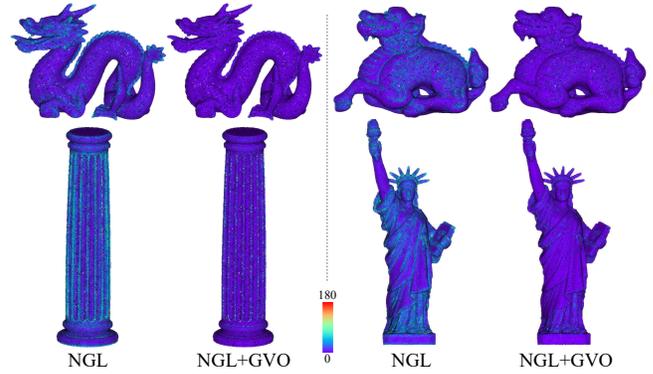

**Figure 4: Error maps of oriented normals from different stages of our method, namely NGL and NGL+GVO. NGL can solve initial normals with consistent orientation but possibly moderate accuracy, and GVO can further improve the accuracy of initial normals to a higher level.**

## 3 COMPARISON WITH SURFACE RECONSTRUCTION METHODS

In our experiments, we have shown the results of utilizing a Poisson reconstruction algorithm [Kazhdan and Hoppe 2013] to reconstruct surfaces from the estimated normals. To further evaluate the reconstructed surfaces, we compare our method with baseline algorithms that are specifically designed for surface reconstruction, including IGR [Gropp et al. 2020], SAP [Peng et al. 2021], SAL [Atzmon and Lipman 2020], Neural-Pull (NP) [Ma et al. 2021], OSP [Ma et al. 2022a] and PCP [Ma et al. 2022b]. As shown in Fig. 7 and Fig. 14, we employ these methods to reconstruct surfaces from point clouds with different noise levels. Based on the accurate normals estimated by our method, the Poisson algorithm can reconstruct more complete detailed geometry than baseline methods from various point cloud data.

## 4 MORE DISCUSSIONS

### 4.1 Discussion on technical contribution and novelty

The two-stage architectures of existing oriented normal estimation paradigms combine two independent algorithms, *i.e.*, unoriented normal estimation and normal orientation, such as PCA+MST. They require a lot of work to tune the parameters of the two algorithms. Our observation is that higher-precision unoriented normals do not necessarily result in more accurate oriented normals using a normal orientation algorithm based on propagation strategy (see Table 1 of the paper). This means that even if we develop better unoriented normal estimation algorithms, utilizing existing normal orientation algorithms will not lead to better orientation results.

Different from most previous works, which first infer unoriented normals based on local geometry and then try to find a consistent orientation of these normals, our proposed approach first learns a signed distance field that provides noisy but correctly-oriented normals (the Neural Gradient Learning module), and then fitting these approximate normals to local surface features by learning an

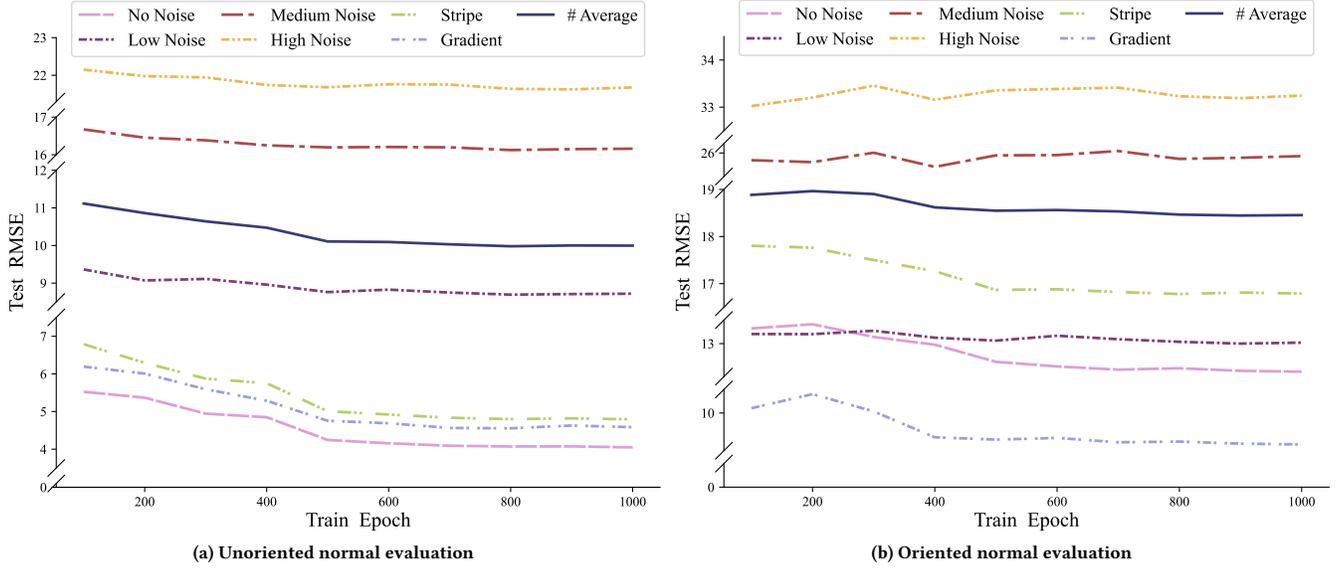

(a) Unoriented normal evaluation        (b) Oriented normal evaluation

**Figure 5: Unoriented and oriented normal evaluation results on the PCPNet test set using network models trained for 100 to 1000 epochs. We plot the results for different noise levels and density variations, and plot the average values of all categories. The estimated normals are evaluated using metrics $\mathrm{RMSE}_U$ and $\mathrm{RMSE}_O$.**

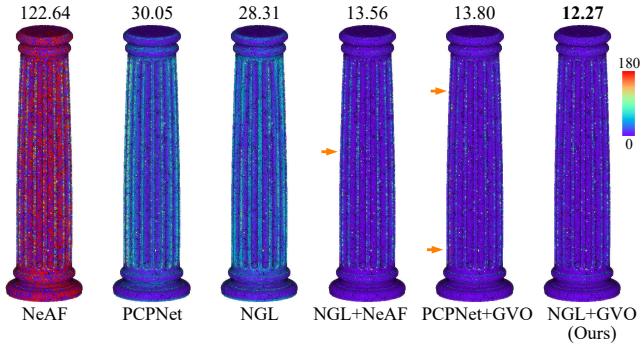

**Figure 6: We integrate our NGL and GVO into other methods to estimate oriented normals. Average values of RMSE are provided. Note that NeAF can not estimate oriented normals. Please zoom in to see the difference.**

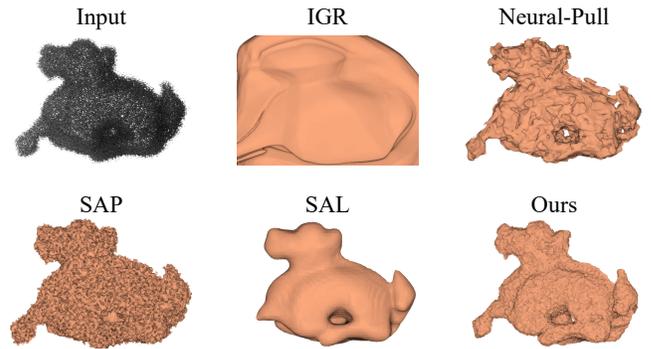

**Figure 7: Comparison with the implicit representation methods in surface reconstruction task. Baseline methods include IGR [Gropp et al. 2020], SAP [Peng et al. 2021], SAL [Atzmon and Lipman 2020] and Neural-Pull [Ma et al. 2021].**

angular distance field from weighted features of a neighborhood of each point cloud (the Gradient Vector Optimization module). Our method achieves the state-of-the-art performance in both unoriented and oriented normal estimation of point clouds.

## 4.2 The differences between the proposed method and SHS-Net

Concurrent work SHS-Net [Li et al. 2023a], published at CVPR 2023, estimates normal in a different way than ours. SHS-Net utilizes the learning of signed hyper surface in the feature space to directly regress oriented normals. It extracts features from local patches and global sampling points in parallel and fuses the two features. In particular, in order to determine the normal orientation of a

query point, its corresponding global sampling point set needs to be found from the entire point cloud. This is time-consuming, especially when dealing with large-scale point cloud data.

In contrast, our method is to first use the implicit representation and the signed distance field to directly learn the coarse correctly-oriented normal from the entire point cloud, and then use the local information to refine the coarse normal and improve the accuracy of direction based on the angular distance field. Our method does not need to perform the global sampling for the query point like SHS-Net dose.

As shown in Tables 1 and 2 of the main paper, our method has the best performance for oriented normal estimation, and has comparable performance for unoriented normal estimation compared

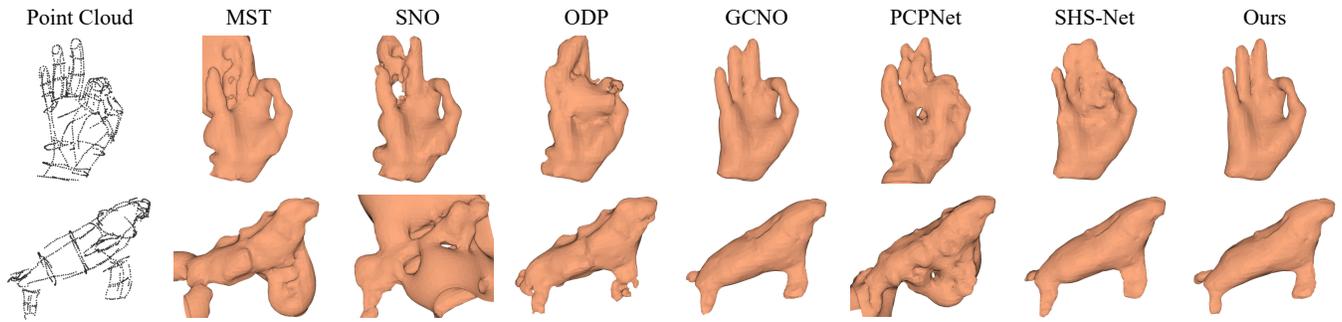

Point Cloud    MST    SNO    ODP    GCNO    PCPNet    SHS-Net    Ours

**Figure 8: Oriented normal estimation for wireframe-type point clouds. MST, SNO and ODP use the unoriented normals estimated by the PCA algorithm as input. Surfaces are reconstructed using the Poisson reconstruction algorithm based on the final estimated normals, and the ground-truth is not available.**

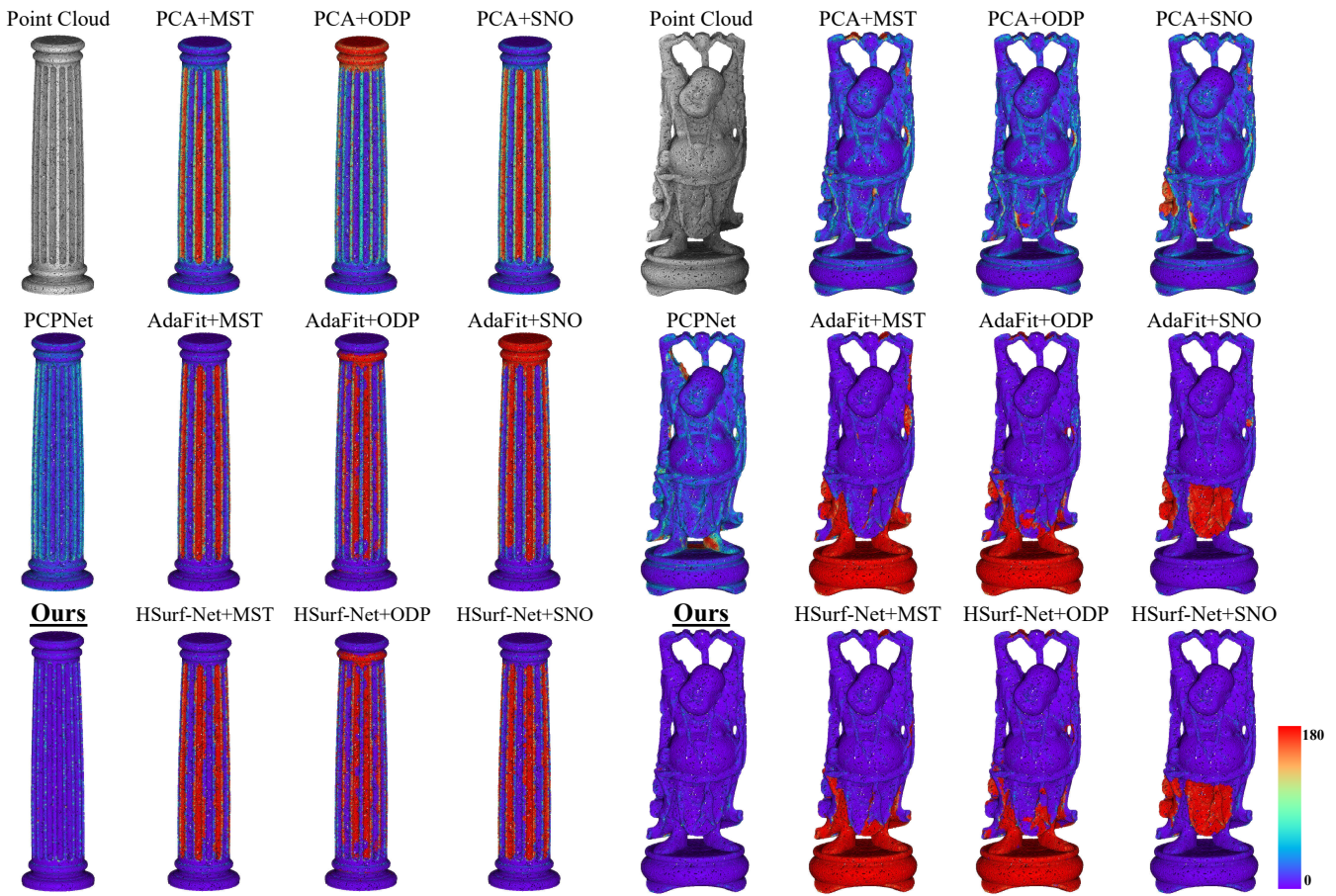

**Figure 9: Visual comparison of oriented normal errors. We map RMSEs to a heatmap, where red (180°) indicates the opposite direction.**

with the method SHS-Net. The results in Table 4 also show that our method has a clear advantage over SHS-Net on sparse point clouds.

## 5 MORE RESULTS

**Sparse data and comparison with GCNO.** The running time of GCNO [Xu et al. 2023] increases so dramatically with the number of points in the point cloud that we were unable to fully test it on existing benchmark datasets. We conduct an evaluation on a

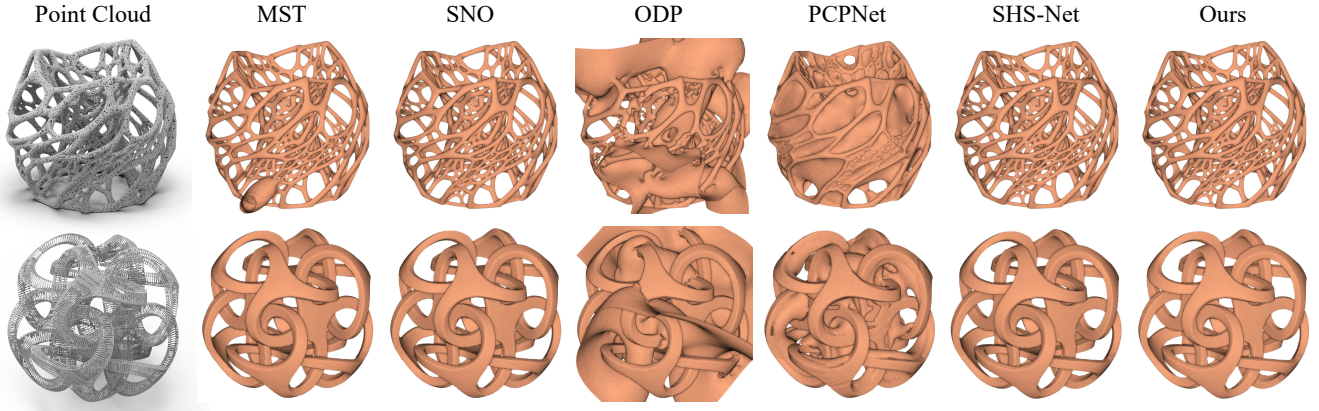

**Figure 10: Oriented normal estimation for point clouds with complex topology. MST, SNO and ODP use the unoriented normals estimated by the PCA algorithm as input. Surfaces are reconstructed using the Poisson reconstruction algorithm based on the final estimated normals.**

dataset that has the same shapes as the FamousShape dataset but each shape in this dataset contains only 5000 points. We report quantitative comparison results of oriented normal estimation in Table 4. The traditional algorithms, such as GCNO and PCA+MST, are implemented in C++ on the Windows platform and run on an Intel i9-11900K CPU. The deep learning-based methods, such as PCPNet, SHS-Net, HSurf-Net+ODP and ours, are implemented in Python on the Ubuntu platform and run on an NVIDIA 2080 Ti GPU. The time is the average number of seconds to process a point cloud with 5000 points, and GCNO takes much longer to run than other methods. Our method has the lowest RMSE result, and this evaluation also demonstrates the good performance of our method on sparse point clouds. Additionally, we provide a comparison on wireframe-type point clouds, which are extremely sparse and unevenly sampled. The experimental results in Fig. 8 show that our method is able to handle wireframe-type data.

**Shape data.** As shown in Fig. 10, we provide two nest-like point clouds with complex topology and geometry. The shape in the top row has 100K points while the shape in the bottom row has 70K points. The results show that our method can handle data with complex topology and accurately estimate oriented normals. In Fig. 9 and Fig. 16, we provide visual comparisons of the oriented normal RMSE on datasets PCPNet and FamousShape, and the results show the clear advantage of our method. We render the point clouds with RGB colors generated from the error values. In Fig. 11, we show the oriented normal PGP curves on different data categories of the PCPNet dataset. It can be seen that our method achieves significant performance improvements at most of the thresholds compared to the baseline methods. In Fig. 17, we show the unoriented normal PGP curves of various methods on datasets PCPNet and FamousShape. We can see that our method has a performance advantage at the vast majority of thresholds. These results demonstrate that our method has clear advantages over existing methods on the task of oriented normal estimation.

**Paris-rue-Madame dataset.** As shown in Fig. 13, we provide more surface reconstruction results on the Paris-rue-Madame dataset [Serna et al. 2014], which is acquired from real-world street scenes using laser scanners. We show a top view of the scene and compare the

reconstructed cars on the street. We can see that the reconstruction algorithm can benefit from the oriented normals estimated by our method to generate better car shapes. The evaluation results on the real-world datasets, including SceneNN and Paris-rue-Madame, demonstrate the good generalization ability of our method.

**SceneNN dataset.** To test the generalization ability of our method, we also evaluate our methods on the SceneNN dataset [Hua et al. 2016], which is acquired from real-world room scenes using RGB-D cameras. As shown in Fig. 15, we visualize the surface reconstruction results based on the oriented normals estimated by different methods. The results show that our method gives better shapes of objects in the kitchen room.

## 6 LIMITATION AND FAILURE CASES

The task of estimating oriented normals with consistent orientations is more challenging than finding the perpendicular lines of planes/surfaces, *i.e.*, unoriented normals. As we introduce in the paper, our method determines the normal orientation based on the gradients learned by NGL. In our experiments, we observe that our GVO can always deliver good normal results from various point clouds in unoriented normal evaluation. However, our NGL may fail to learn the normal orientations for oriented normal estimation. This will lead to poor evaluation results for oriented normals even though their corresponding unoriented normals are very accurate. As encountered in existing oriented normal estimation methods, the bottleneck of estimating high-precision oriented normals is correctly determining the normal orientation. As shown in Fig. 12, we provide some failure cases of our method in oriented normal estimation. In these cases, the unoriented normals are accurate, and the normal orientations of most points are also correct.

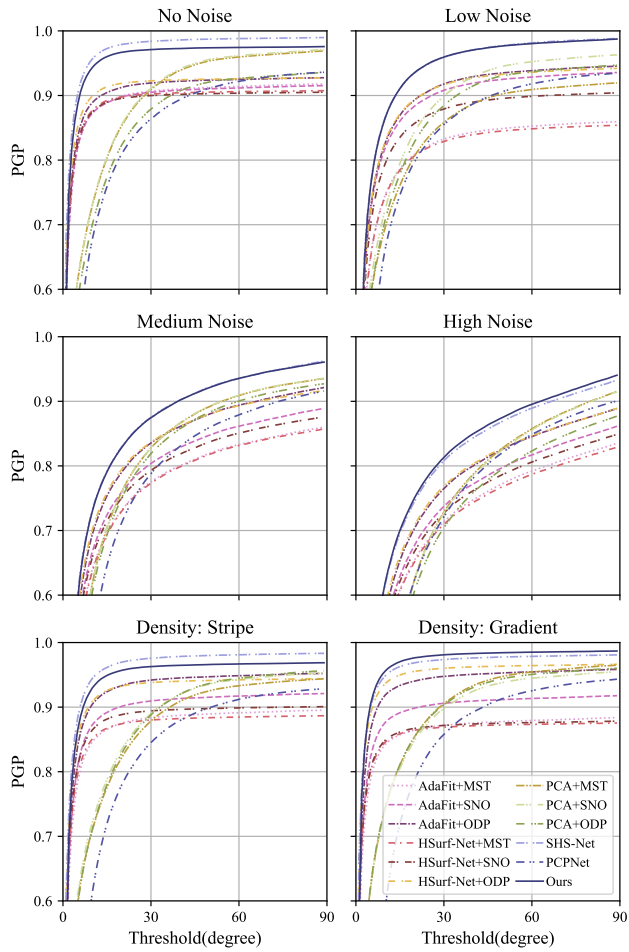

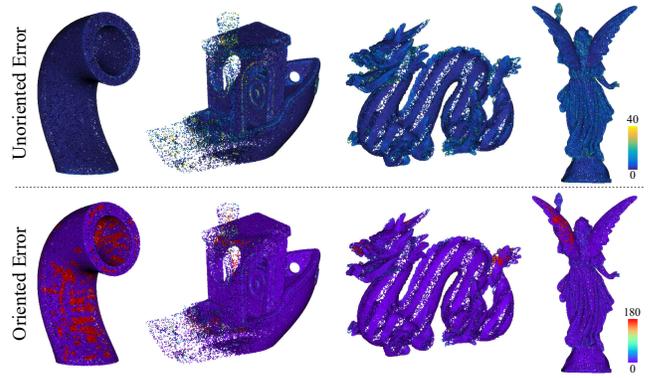

**Figure 12: Visualization of some failure cases. Our method can estimate unoriented normals with high accuracy (top row). However, the normal orientations, *i.e.*, gradients from NGL, in some local areas are not correctly learned (bottom row).** $\text{RMSE}_U$ **and** $\text{RMSE}_O$ **of point cloud normals are mapped to different heatmaps for visualization.**

**Figure 11: Oriented normal PGP curves on the PCPNet dataset. Our method achieves significant performance improvements at most of the thresholds. The X-axis is the angle threshold and the Y-axis is the percentage of good point normals (PGP) whose errors are less than the given threshold.**

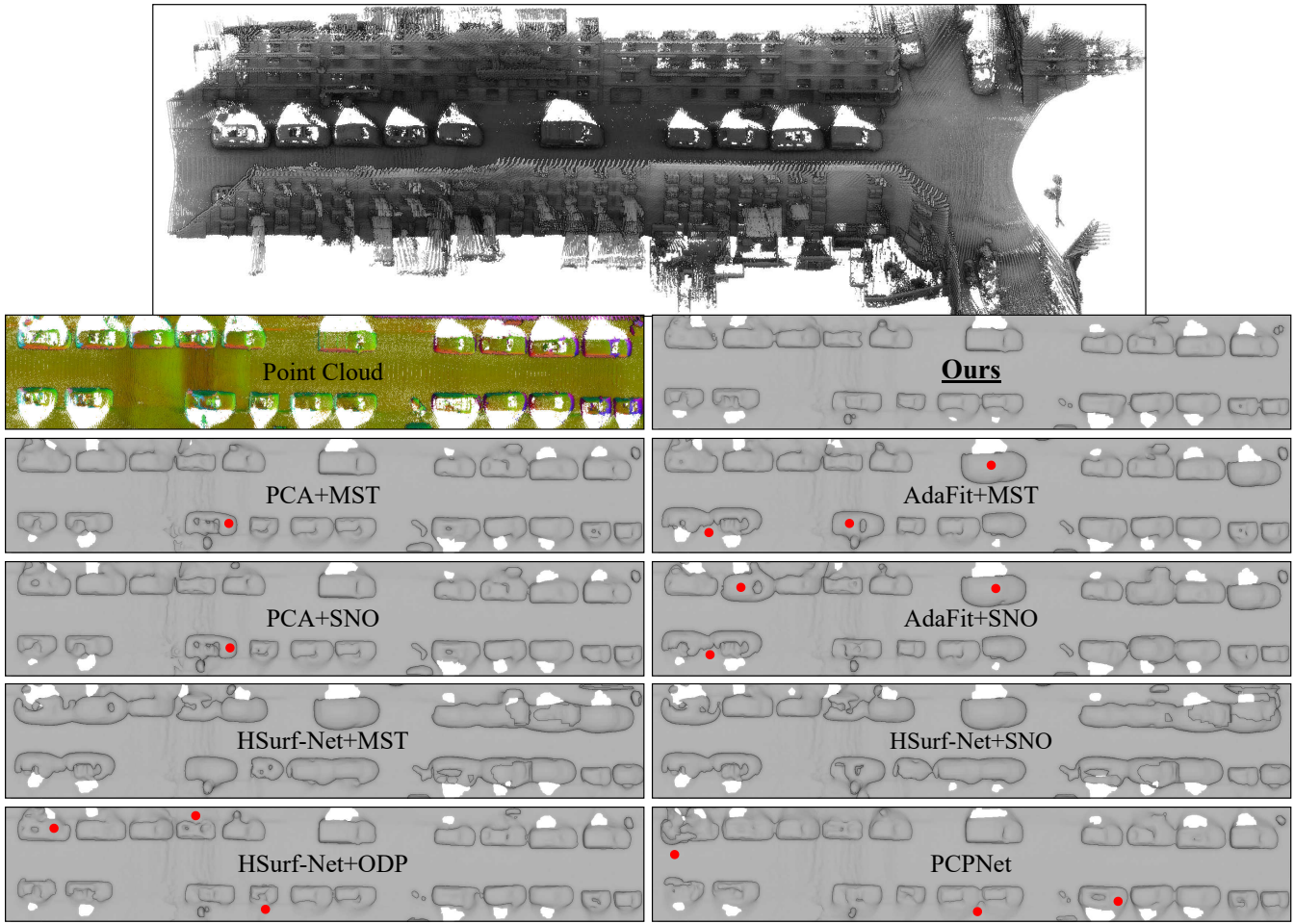

**Figure 13: The reconstructed surfaces using oriented normals estimated by different methods on the Paris-rue-Madame dataset. Our method can provide more complete and clear car shapes. We mark those less obvious differences with red dots to help distinguish them. The colored point cloud is obtained by coloring it with our estimated normals.**

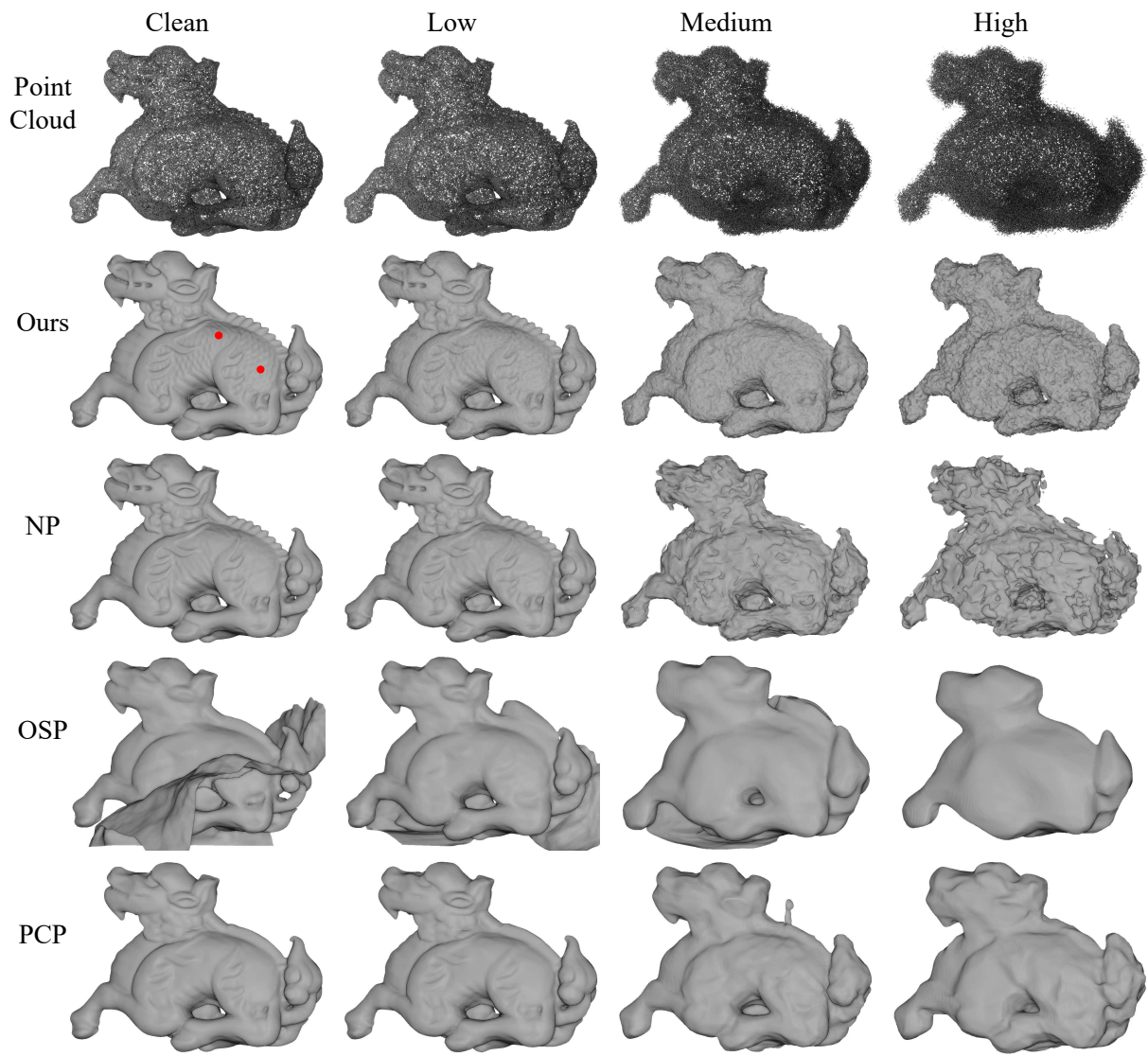

**Figure 14: Comparison with surface reconstruction methods that use point pulling strategy, including Neural-Pull (NP) [Ma et al. 2021], OSP [Ma et al. 2022a] and PCP [Ma et al. 2022b], on point clouds with different noise levels.**

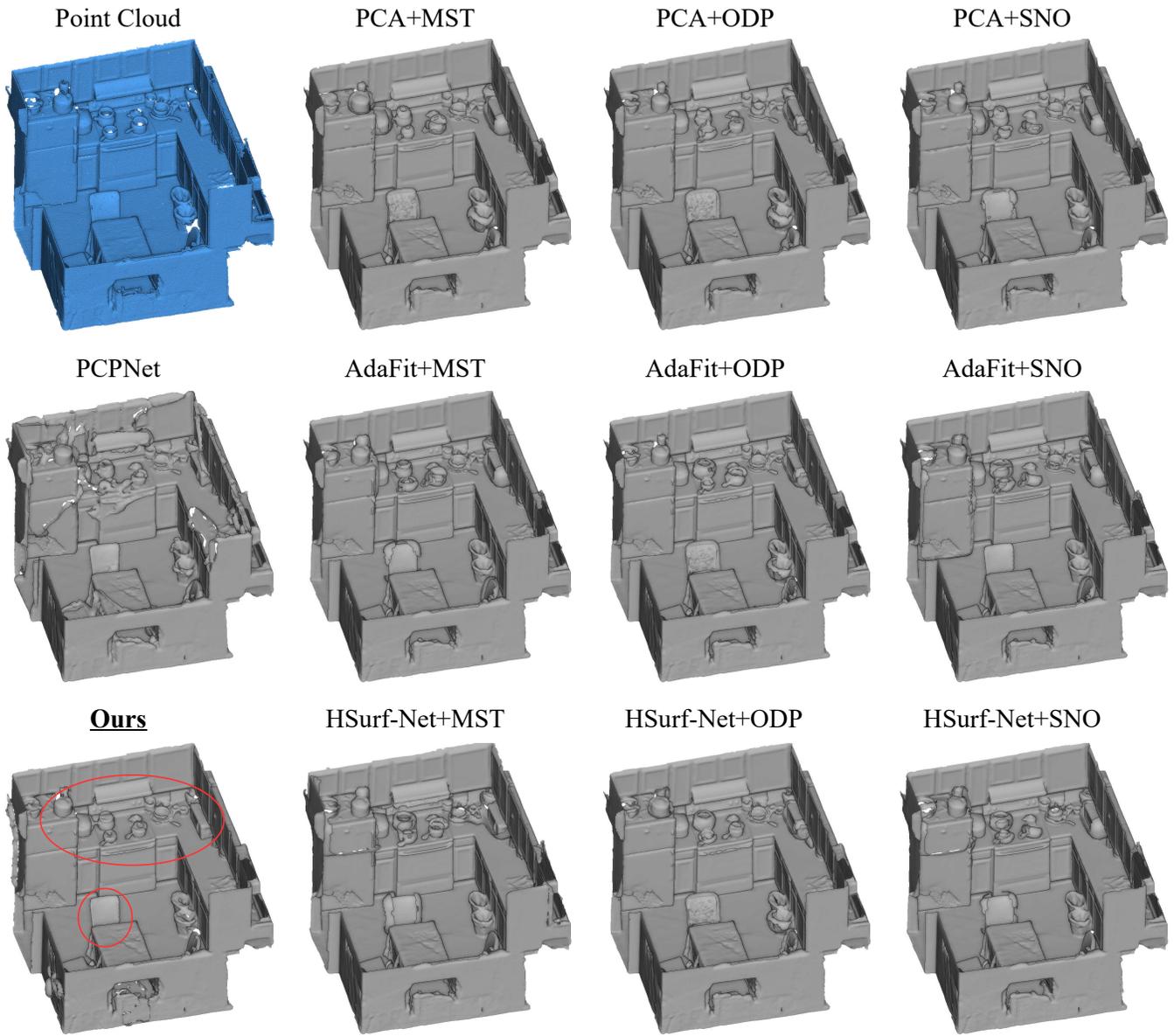

**Figure 15: The reconstructed surfaces using oriented normals estimated by different methods on the SceneNN dataset. The differences are mainly concentrated in the area inside the red circle.**

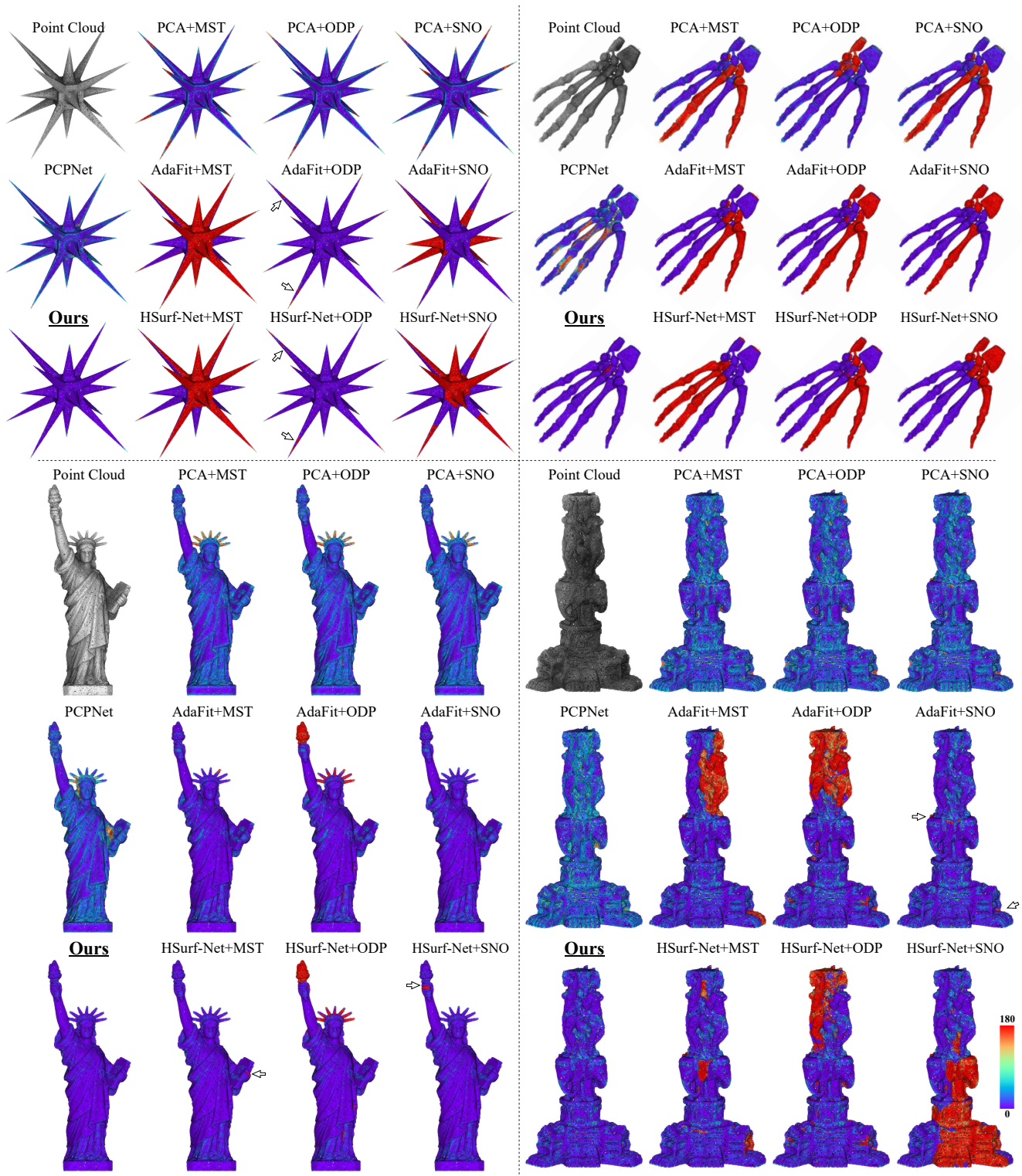

**Figure 16:** Visualization of the oriented normal error on datasets PCPNet (left) and FamousShape (right). The angle error is mapped to a heatmap ranging from $0°$ to $180°$. The purple color indicates the same direction as the ground-truth, while the red color is the opposite.

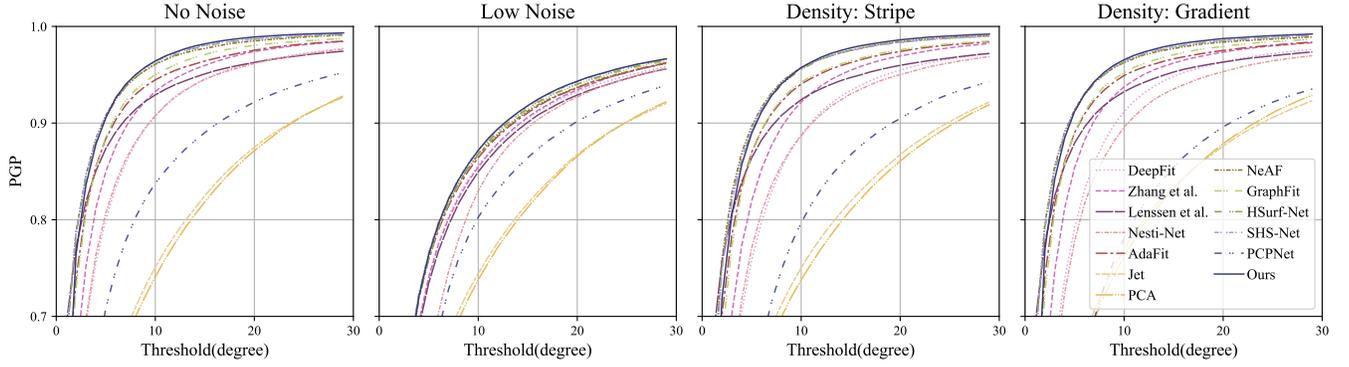

(a) Unoriented normal PGP curves on the PCPNet dataset.

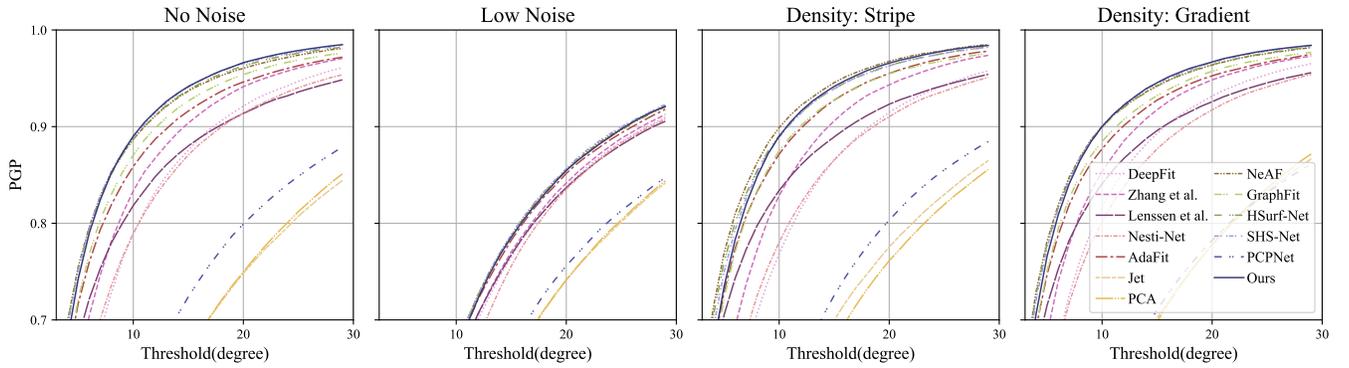

(b) Unoriented normal PGP curves on the FamousShape dataset.

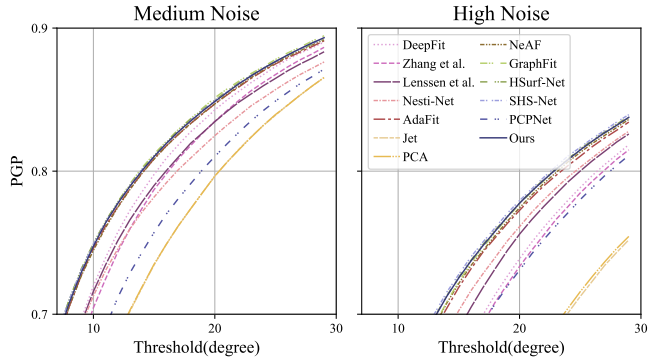

(c) Unoriented normal PGP curves on the PCPNet dataset.

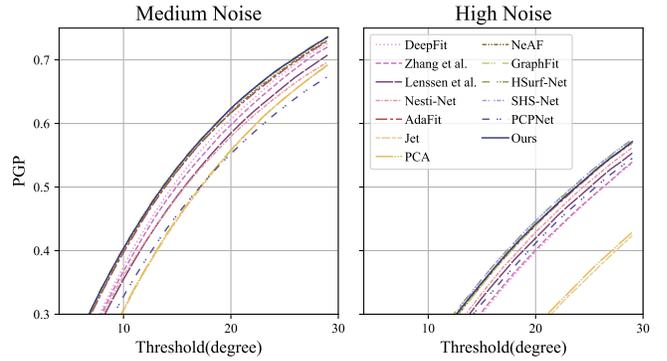

(d) Unoriented normal PGP curves on the FamousShape dataset.

Figure 17: Unoriented normal PGP curves of different methods on datasets PCPNet (a)(c) and FamousShape (b)(d). Our method maintains a performance advantage at the vast majority of thresholds. The axis scale of (a) and (b) is different from that of (c) and (d) as we enlarge it in (c) and (d) for better presentation. The X-axis is the angle threshold and the Y-axis is the percentage of good point normals (PGP) whose errors are less than the given threshold.



\end{document}

%% file: table_pcpnet_famousShape_orient.tex
\begin{table*}[t]
\centering
%\small
\footnotesize
\setlength{\tabcolsep}{1.75mm}
\caption{
	RMSE of oriented normals on datasets PCPNet and FamousShape.
	$\ast$ means the source code is uncompleted.
}
\vspace{-0.35cm}
\label{table:pcpnet_famousShape_o}
% \resizebox{\linewidth}{!}{
\begin{tabular}{l|cccc|cc| >{\columncolor{mygray}} c||  cccc|cc| >{\columncolor{mygray}} c}
	\toprule
	\multirow{3}{*}{Category} & \multicolumn{7}{c||}{\textbf{PCPNet Dataset}} & \multicolumn{7}{c}{\textbf{FamousShape Dataset}} \\
	\cmidrule(r){2-15}
	& \multicolumn{4}{c|}{Noise} & \multicolumn{2}{c|}{Density} &     & \multicolumn{4}{c|}{Noise} & \multicolumn{2}{c|}{Density} &    \\
	& None & 0.12\% & 0.6\% & 1.2\%  & Stripe & Gradient  & \multirow{-2}{*}{{Average}} & None & 0.12\% & 0.6\% & 1.2\%  & Stripe & Gradient & \multirow{-2}{*}{{Average}} \\
	\midrule
	PCA+MST~\cite{hoppe1992surface}
	& 19.05 & 30.20 & 31.76 & 39.64 & 27.11 & 23.38 &   28.52    & 35.88 & 41.67 & \textbf{38.09} & 60.16 & 31.69 & 35.40 &   40.48  \\
	PCA+SNO~\cite{schertler2017towards}
	& 18.55 & 21.61 & 30.94 & 39.54 & 23.00 & 25.46 &   26.52    & 32.25 & 39.39 & 41.80 & 61.91 & 36.69 & 35.82 &   41.31  \\
	PCA+ODP~\cite{metzer2021orienting}
	& 28.96 & 25.86 & 34.91 & 51.52 & 28.70 & 23.00 &   32.16    & 30.47 & 31.29 & 41.65 & 84.00 & 39.41 & 30.72 &   42.92  \\
	AdaFit~\cite{zhu2021adafit}+MST
	& 27.67 & 43.69 & 48.83 & 54.39 & 36.18 & 40.46 &   41.87    & 43.12 & 39.33 & 62.28 & 60.27 & 45.57 & 42.00 &   48.76  \\
	AdaFit~\cite{zhu2021adafit}+SNO
	& 26.41 & 24.17 & 40.31 & 48.76 & 27.74 & 31.56 &   33.16    & 27.55 & 37.60 & 69.56 & 62.77 & 27.86 & 29.19 &   42.42  \\
	AdaFit~\cite{zhu2021adafit}+ODP
	& 26.37 & 24.86 & 35.44 & 51.88 & 26.45 & 20.57 &   30.93    & 41.75 & 39.19 & 44.31 & 72.91 & 45.09 & 42.37 &   47.60  \\
	HSurf-Net~\cite{li2022hsurf}+MST
	& 29.82 & 44.49 & 50.47 & 55.47 & 40.54 & 43.15 &   43.99    & 54.02 & 42.67 & 68.37 & 65.91 & 52.52 & 53.96 &   56.24  \\
	HSurf-Net~\cite{li2022hsurf}+SNO
	& 30.34 & 32.34 & 44.08 & 51.71 & 33.46 & 40.49 &   38.74    & 41.62 & 41.06 & 67.41 & 62.04 & 45.59 & 43.83 &   50.26  \\
	HSurf-Net~\cite{li2022hsurf}+ODP
	& 26.91 & 24.85 & 35.87 & 51.75 & 26.91 & 20.16 &   31.07    & 43.77 & 43.74 & 46.91 & 72.70 & 45.09 & 43.98 &   49.37  \\
	PCPNet~\cite{guerrero2018pcpnet}
	& 33.34 & 34.22 & 40.54 & 44.46 & 37.95 & 35.44 &   37.66    & 40.51 & 41.09 & 46.67 & 54.36 & 40.54 & 44.26 &   44.57  \\
	DPGO$^\ast$~\cite{wang2022deep}
	& 23.79 & 25.19 & 35.66 & 43.89 & 28.99 & 29.33 &   31.14    & - & - & - & - & - & - & -  \\
	SHS-Net~\cite{li2023shsnet}
	& \textbf{10.28} & {13.23} & \textbf{25.40} & 35.51 & \textbf{16.40} & 17.92 &   19.79    & 21.63 & 25.96 & 41.14 & 52.67 & \textbf{26.39} & 28.97 &   32.79  \\
	Ours
	&        {12.52} & \textbf{12.97} & {25.94} & \textbf{33.25} & {16.81} & \textbf{9.47} &   \textbf{18.49}
	& \textbf{13.22} & \textbf{18.66} & {39.70} & \textbf{51.96} & {31.32} & \textbf{11.30} &   \textbf{27.69}  \\
	\bottomrule
\end{tabular} %}
% \vspace{-0.1cm}
\end{table*}

%% file: table_pcpnet_famousShape_unorient.tex
\begin{table*}[t]
\centering
%\small
\footnotesize
\setlength{\tabcolsep}{1.6mm}
\caption{
	RMSE of unoriented normal on datasets PCPNet and FamousShape.
	% Sorted by the average values on the PCPNet dataset.
	% Lower is better.
	$\ast$ means the source code is uncompleted or unavailable.
	% we show the results for varying levels of noise, from zero noise to high noise.
	% The two rows in the middle show the results for point clouds with a non-uniform sampling rate.
	% The last row shows the global average error over all shapes.
}
\vspace{-0.35cm}
\label{table:pcpnet_famousShape}
% \resizebox{.95\linewidth}{!}{
\begin{tabular}{l|cccc|cc| >{\columncolor{mygray}} c||  cccc|cc| >{\columncolor{mygray}} c}
	\toprule
	\multirow{3}{*}{Category} & \multicolumn{7}{c||}{\textbf{PCPNet Dataset}} & \multicolumn{7}{c}{\textbf{FamousShape Dataset}} \\
	\cmidrule(r){2-15}
	% \cline{2-15}
	% \rowcolor{white}
	& \multicolumn{4}{c|}{Noise} & \multicolumn{2}{c|}{Density} &     & \multicolumn{4}{c|}{Noise} & \multicolumn{2}{c|}{Density} &    \\
	% \rowcolor{white}
	& None & 0.12\% & 0.6\% & 1.2\%  & Stripe & Gradient  & \multirow{-2}{*}{{Average}} & None & 0.12\% & 0.6\% & 1.2\%  & Stripe & Gradient & \multirow{-2}{*}{{Average}} \\
	\midrule
	Jet~\cite{cazals2005estimating}              & 12.35 & 12.84 & 18.33 & 27.68 & 13.39 & 13.13 &  16.29    & 20.11 & 20.57 & 31.34 & 45.19 & 18.82 & 18.69 &   25.79  \\
	PCA~\cite{hoppe1992surface} 	             & 12.29 & 12.87 & 18.38 & 27.52 & 13.66 & 12.81 &  16.25    & 19.90 & 20.60 & 31.33 & 45.00 & 19.84 & 18.54 &   25.87  \\
	% HoughCNN~\cite{boulch2016deep}             & 10.23 & 11.62 & 22.66 & 33.39 & 11.02 & 12.47 &  16.90
	PCPNet~\cite{guerrero2018pcpnet}             & 9.64  & 11.51 & 18.27 & 22.84 & 11.73 & 13.46 &  14.58    & 18.47 & 21.07 & 32.60 & 39.93 & 18.14 & 19.50 &   24.95  \\
	Zhou \etal$^\ast$~\cite{zhou2020normal}      & 8.67  & 10.49 & 17.62 & 24.14 & 10.29 & 10.66 &  13.62    & -    & -     & -    & -    & -    & -    & -    \\
	Nesti-Net~\cite{ben2019nesti}                & 7.06  & 10.24 & 17.77 & 22.31 & 8.64  & 8.95  &  12.49    & 11.60 & 16.80 & 31.61 & 39.22 & 12.33 & 11.77 &   20.55  \\
	Lenssen \etal~\cite{lenssen2020deep}         & 6.72  & 9.95  & 17.18 & 21.96 & 7.73  & 7.51  &  11.84    & 11.62 & 16.97 & 30.62 & 39.43 & 11.21 & 10.76 &   20.10  \\
	DeepFit~\cite{ben2020deepfit}                & 6.51  & 9.21  & 16.73 & 23.12 & 7.92  & 7.31  &  11.80    & 11.21 & 16.39 & 29.84 & 39.95 & 11.84 & 10.54 &   19.96  \\
	MTRNet$^\ast$~\cite{cao2021latent}           & 6.43  & 9.69  & 17.08 & 22.23 & 8.39  & 6.89  &  11.78    & -    & -     & -    & -    & -    & -    & -    \\
	Refine-Net~\cite{zhou2022refine}             & 5.92  & 9.04  & 16.52 & 22.19 & 7.70  & 7.20  &  11.43    & -    & -     & -    & -    & -    & -    & -    \\
	Zhang \etal$^\ast$~\cite{zhang2022geometry}  & 5.65  & 9.19  & 16.78 & 22.93 & 6.68  & 6.29  &  11.25    & 9.83 & 16.13 & 29.81 & 39.81 & 9.72 & 9.19 &   19.08  \\
	Zhou \etal$^\ast$~\cite{zhou2023improvement} & 5.90  & 9.10  & 16.50 & 22.08 & 6.79  & 6.40  &  11.13    & -    & -     & -    & -    & -    & -    & -    \\
	AdaFit~\cite{zhu2021adafit}                  & 5.19  & 9.05  & 16.45 & 21.94 & 6.01  & 5.90  &  10.76    & 9.09 & 15.78 & 29.78 & 38.74 & 8.52 & 8.57 &   18.41  \\
	GraphFit~\cite{li2022graphfit}               & 5.21  & 8.96  & \textbf{16.12} & 21.71 & 6.30  & 5.86  &  10.69    & 8.91 & 15.73 & 29.37 & 38.67 & 9.10 & 8.62 &   18.40  \\
	NeAF~\cite{li2023NeAF}                       & 4.20  & 9.25  & 16.35 & 21.74 & 4.89  & 4.88  &  10.22    & 7.67 & 15.67 & 29.75 & 38.76 & \textbf{7.22} & 7.47 &   17.76  \\
	HSurf-Net~\cite{li2022hsurf}                 & 4.17  & 8.78  & 16.25 & 21.61 & 4.98  & 4.86  &  10.11    & 7.59 & 15.64 & 29.43 & \textbf{38.54} & 7.63 & 7.40 &   17.70  \\
	SHS-Net~\cite{li2023shsnet}  	             & \textbf{3.95}  & \textbf{8.55}  & 16.13 & \textbf{21.53} & 4.91  & 4.67  &  \textbf{9.96}
												 & 7.41  & \textbf{15.34} & \textbf{29.33} & 38.56 & 7.74 & 7.28 &   \textbf{17.61}  \\
	Ours  	&        {4.06} & {8.70}  & \textbf{16.12} & {21.65} & \textbf{4.80} & \textbf{4.56} & \textbf{9.98}
			& \textbf{7.25} & {15.60} & \textbf{29.35} & {38.74} &        {7.60} & \textbf{7.20} & \textbf{17.62}   \\
	\bottomrule
\end{tabular} %}
% \vspace{-0.1cm}
\end{table*}

%% file: table_time.tex
\begin{table}[t]
    % \small
    \footnotesize
    \centering
    \setlength{\tabcolsep}{1.8mm}
    \caption{
      Comparison of the RMSE, number of learnable network parameters (million), and test runtime (seconds per 100k points) for learning-based oriented normal estimation methods.
    }
    \vspace{-0.35cm}
    \begin{tabular}{lccccc}
    \toprule
     %  &  Ours   & \tabincell{c}{PCPNet\\ \cite{guerrero2018pcpnet}}
     %            & \tabincell{c}{HSurf-Net~\cite{li2022hsurf}\\+ODP~\cite{metzer2021orienting}}
     %            & \tabincell{c}{AdaFit~\cite{zhu2021adafit}\\+ODP}    \\
                  & HSurf-Net+ODP  & AdaFit+ODP  & PCPNet          & SHS-Net  &  Ours            \\
    \midrule
    RMSE          & 31.07          & 30.93       & 37.66           & 19.79    & \textbf{18.49}   \\
    Param.        & 2.59           & 5.30        & 22.36           & 3.27     & \textbf{2.38}    \\
    Time          & 308.82         & 304.77      & \textbf{63.02}  & 65.89    &        {71.29}   \\
    \bottomrule
    \end{tabular}
    \label{tab:time}
    % \vspace{-0.3cm}
 \end{table}

%% file: table_ablation.tex
\begin{table*}[t]
\centering
% \small
\footnotesize
\setlength{\tabcolsep}{1.95mm}
\caption{
    Ablation studies with the metric of unoriented and oriented normal on the PCPNet dataset.
    Please see the text for more details.
    % See the text for discussion.
}
\vspace{-0.35cm}
\label{tab:ablation}
% \resizebox{\linewidth}{!}{
\begin{tabular}{ll|cccc|cc| >{\columncolor{mygray}} c||  cccc|cc| >{\columncolor{mygray}} c}
    \toprule
    & \multirow{3}{*}{Category} & \multicolumn{7}{c||}{\textbf{Unoriented Normal}} & \multicolumn{7}{c}{\textbf{Oriented Normal}} \\
    \cmidrule(r){3-16}
    &
    & \multicolumn{4}{c|}{Noise} & \multicolumn{2}{c|}{Density} &    & \multicolumn{4}{c|}{Noise} & \multicolumn{2}{c|}{Density} &  \\
    &
    & None & 0.12\% & 0.6\% & 1.2\% & Stripe & Gradient & \multirow{-2}{*}{{Average}}
    & None & 0.12\% & 0.6\% & 1.2\% & Stripe & Gradient & \multirow{-2}{*}{{Average}} \\
    \midrule
    \multirow{4}{*}{\textbf{(a)}}
    & w/o NGL
    & 4.20 & 8.78 & 16.16 & 21.67 & 4.88 & 4.64 & 10.06 & 124.53 & 123.11 & 120.35 & 117.44 & 123.57 & 118.80 & 121.30  \\
    & w/o GVO
    & 12.24 & 12.74 & 17.89 & 23.88 & 15.16 & 13.75 & 15.94 & 18.39 & 15.32 & 25.20 & 32.57 & 22.91 & 15.73 & 21.69  \\
    & w/o inlier score
    & 4.26 & 8.94 & 16.11 & 21.70 & 5.26 & 5.00 & 10.21 & 12.78 & 13.25 & 25.99 & 33.43 & 17.30 & 9.82 & 18.76  \\
    & w/o $w$ in kernel
    & 4.11 & 8.71 & 16.14 & 21.63 & 5.11 & 4.80 & 10.08 & 12.38 & 12.94 & 25.88 & 33.30 & 16.87 & 9.47 & 18.47  \\
    \hline
    \multirow{4}{*}{\textbf{(b)}}
    % & $\mathcal{L}_2$ w/ MSE
    % & 4.55 & 8.88 & 16.17 & 21.41 & 5.48 & 5.08 & 10.26 & 12.84 & 13.18 & 25.95 & 33.27 & 17.23 & 9.84 & 18.72  \\
    & $\mathcal{L}_{\sss \text{NGL}}$ (L1)
    & 4.09 & 8.69 & 16.13 & 21.65 & 4.80 & 4.57 & 9.99 & 17.27 & 12.27 & 35.58 & 37.95 & 11.26 & 9.28 & 20.60  \\
    & $\mathcal{L}_{\sss \text{NGL}}$ (MSE)
    & 4.08 & 8.70 & 16.13 & 21.64 & 4.82 & 4.58 & 9.99 & 21.71 & 18.82 & 27.81 & 33.38 & 13.29 & 11.68 & 21.12  \\
    & $\mathcal{L}_{\sss \text{GVO}}$($\lambda \!=\! 0.2$)
    & 4.12 & 8.75 & 16.16 & 21.74 & 5.09 & 4.71 & 10.10 & 12.60 & 12.99 & 25.98 & 33.34 & 16.90 & 9.57 & 18.56  \\
    & $\mathcal{L}_{\sss \text{GVO}}$($\lambda \!=\! 0.8$)
    & 4.14 & 8.82 & 16.18 & 21.64 & 4.96 & 4.74 & 10.08 & 12.58 & 13.09 & 26.04 & 33.33 & 16.87 & 9.45 & 18.56  \\
    \hline
    \multirow{3}{*}{\textbf{(c)}}
    & $k \!=\! 1$
    & 4.07 & 8.70 & 16.13 & 21.65 & 4.79 & 4.55 & 9.98 & 13.57 & 18.24 & 38.29 & 47.23 & 9.27 & 8.99 & 22.60  \\
    & $k \!=\! 32$
    & 4.06 & 8.69 & 16.13 & 21.65 & 4.79 & 4.56 & 9.98 & 13.64 & 24.31 & 29.83 & 33.93 & 17.37 & 8.51 & 21.27  \\
    & $k \!=\! 128$
    & 4.08 & 8.70 & 16.13 & 21.64 & 4.84 & 4.58 & 9.99 & 12.84 & 23.65 & 34.96 & 33.03 & 37.64 & 18.42 & 26.76  \\
    \hline
    \multirow{2}{*}{\textbf{(d)}}
    & $d_{\sigma} \!=\! 32$th
    & 4.07 & 8.69 & 16.12 & 21.66 & 4.83 & 4.56 & 9.99 & 12.86 & 23.75 & 29.68 & 36.67 & 10.97 & 8.92 & 20.47  \\
    & $d_{\sigma} \!=\! 64$th
    & 4.08 & 8.70 & 16.13 & 21.64 & 4.81 & 4.57 & 9.99 & 13.77 & 18.98 & 29.84 & 33.25 & 18.41 & 8.87 & 20.52  \\
    \hline
    \multirow{4}{*}{\textbf{(e)}}
    % & uniform $\mathcal{D'}$
    % & 4.96 & 8.87 & 16.35 & 21.87 & 6.35 & 5.81 & 10.70 & 12.62 & 12.61 & 25.40 & 32.58 & 16.73 & 9.41 & 18.22  \\
    & $\eta \!=\! 0.3$
    & 4.10 & 8.70 & 16.14 & 21.64 & 4.87 & 4.62 & 10.01 & 12.46 & 13.01 & 25.85 & 33.18 & 16.78 & 9.47 & 18.46  \\
    & $\eta \!=\! 0.5$
    & 4.06 & 8.69 & 16.12 & 21.64 & 4.80 & 4.55 & 9.98 & 12.54 & 13.04 & 25.91 & 33.26 & 16.77 & 9.39 & 18.49  \\
    & $M_2 \!=\! 3000$
    & 4.07 & 8.70 & 16.13 & 21.65 & 4.82 & 4.57 & 9.99 & 12.55 & 13.05 & 25.90 & 33.23 & 16.79 & 9.40 & 18.49  \\
    & $M_2 \!=\! 5000$
    & 4.06 & 8.70 & 16.12 & 21.65 & 4.81 & 4.56 & 9.98 & 12.47 & 13.01 & 25.90 & 33.22 & 16.72 & 9.30 & 18.44  \\
    \hline
    & \multirow{1}{*}{\textbf{Full}}
    & {4.06}  & {8.70}  & {16.12} & {21.65} & {4.80}  & {4.56} & {9.98}
    & {12.52} & {12.97} & {25.94} & {33.25} & {16.81} & {9.47} & {18.49}  \\
    \bottomrule
\end{tabular} %}
% \vspace{-0.4cm}
\end{table*}